\documentclass[11pt]{article}
\pdfoutput=1

\usepackage{adjustbox}
\usepackage{algpseudocodex}
\usepackage{alltt}
\usepackage{amsmath, amsfonts, amssymb, mathtools}
\usepackage{amsthm}

\usepackage{array} 
\usepackage{arydshln}
\usepackage{bold-extra}
\usepackage{booktabs}
\usepackage[small]{caption}
\usepackage{color}
\usepackage{comment}
\usepackage{contour}
\usepackage{diagbox}
\usepackage{enumerate}
\usepackage[shortlabels]{enumitem}
\usepackage{epstopdf}
\usepackage{fancyvrb}
\usepackage{float}
\usepackage[bottom]{footmisc}
\usepackage{fvextra}
\usepackage{cellspace}

\usepackage[
  margin=0.83in,
  headheight=12pt,
  headsep=25pt,
  includefoot,
  footskip=30pt,
]{geometry}
\usepackage{graphicx}
\usepackage[final]{hyperref}
\usepackage{inconsolata}
\usepackage[utf8]{inputenc}
\usepackage{listings}
\usepackage{makecell}
\usepackage{makeidx}
\usepackage{mathtools}
\usepackage{mdframed}
\usepackage{multirow}
\usepackage{MnSymbol,wasysym}
\usepackage[numbers]{natbib}
\usepackage{rotating}
\usepackage{scalerel,stackengine}
\usepackage{soul}
\usepackage{subcaption}
\usepackage{tabularx} 
\usepackage[most]{tcolorbox}
\usepackage{thm-restate}
\usepackage{tikz}
\usepackage{pgfplots}
\usepackage{filecontents}
\usepackage{todonotes}
\usepackage{ulem}
\usepackage{url}
\usepackage{xr}
\usepackage{xcolor}
\usepackage{xspace}
\usepackage{algorithm}

\newcommand{\modelwithoutspace}{\textsf{phi-4}}
\newcommand{\model}{\modelwithoutspace~}
\newcommand{\phithree}{\textsf{phi-3}}

\newcommand{\phithreemed}{\textsf{phi-3-medium}}

\definecolor{commentcolour}{rgb}{0.3,0.7,0.2}
\definecolor{backcolour}{rgb}{0.98,0.98,0.98}
\lstdefinelanguage{markdown}{
    comment=[l]{\#},
    morestring=[s]{```}{```},
    commentstyle=\color{commentcolour}\bfseries,
    stringstyle=\color{blue},
    basicstyle=\scriptsize\ttfamily,
    showstringspaces=false,
    breaklines=true,
    breakautoindent=false,
    breakindent=0pt,
    backgroundcolor=\color{backcolour},
}
\lstdefinestyle{mystyle}{
    morekeywords={self},
    basicstyle=\scriptsize\ttfamily,
    keywordstyle=\color{blue},
    commentstyle=\color{commentcolour}\bfseries,
    breaklines=true,
    breakautoindent=false,
    showstringspaces=false,
    backgroundcolor=\color{backcolour},
    stringstyle=\color{red},
}
\lstdefinelanguage{PythonPlus}[]{Python}{
  alsoother={@},
  morekeywords=[1]{,as,assert,nonlocal,with,yield,self,True,False,None} 
  morekeywords=[2]{,__init__,__add__,__mul__,__div__,__sub__,__call__,__getitem__,__setitem__,__eq__,__ne__,__nonzero__,__rmul__,__radd__,__repr__,__str__,__get__,__truediv__,__pow__,__name__,__future__,__all__,}, 
  morekeywords=[3]{,object,type,isinstance,copy,deepcopy,zip,enumerate,reversed,list,set,len,dict,tuple,range,xrange,append,execfile,real,imag,reduce,str,repr,}, 
  morekeywords=[4]{,Exception,NameError,IndexError,SyntaxError,TypeError,ValueError,OverflowError,ZeroDivisionError,}, 
  morekeywords=[5]{,ode,fsolve,sqrt,exp,sin,cos,arctan,arctan2,arccos,pi, array,norm,solve,dot,arange,isscalar,max,sum,flatten,shape,reshape,find,any,all,abs,plot,linspace,legend,quad,polyval,polyfit,hstack,concatenate,vstack,column_stack,empty,zeros,ones,rand,vander,grid,pcolor,eig,eigs,eigvals,svd,qr,tan,det,logspace,roll,min,mean,cumsum,cumprod,diff,vectorize,lstsq,cla,eye,xlabel,ylabel,squeeze,}, 
}
\setuptodonotes{inline}
\usetikzlibrary{matrix}
\usetikzlibrary{shapes,calc,positioning}

\global

\tcbset{
  aibox/.style={
    width=\textwidth,
    top=0pt, bottom=0pt, left=5pt, right=5pt,
    colback=white,
    colframe=black,
    colbacktitle=black,
    enhanced,
    center,
    attach boxed title to top left={yshift=-0.1in,xshift=0.15in},
    boxed title style={boxrule=0pt,colframe=white,},
  }
}
\newtcolorbox{AIbox}[2][]{aibox,title=#2,#1}

\newtcbox{\tokenhilightbox}[1][]{
    on line,
    boxrule=0.5pt,
    colframe=black,
    colback=white,
    boxsep=1.5pt,
    left=0pt,
    right=0pt,
    top=0pt,
    bottom=0pt,
}

\definecolor{aigold}{RGB}{244,210, 1} 
\definecolor{aigreen}{RGB}{210,244,211} 

\sethlcolor{aigreen}

\definecolor{aired}{RGB}{255,180,181}

\newtcbox{\mybox}[1][green]{on line,
arc=0pt,outer arc=0pt,colback=#1!10!white,colframe=#1!50!black,
boxsep=0pt,left=0pt,right=0pt,top=0pt,bottom=0pt,
boxrule=0pt,bottomrule=0pt,toprule=0pt}

\makeatletter
\newcommand{\dynscriptsize}{\check@mathfonts\fontsize{\sf@size}{\z@}\selectfont}
\makeatother
\newcommand\textunderset[2]{%
  \leavevmode
  \vtop{\offinterlineskip
    \halign{%
      \hfil##\hfil\cr 
      \strut#2\cr
      \noalign{\kern-.3ex}
      \dynscriptsize\strut#1\cr
    }%
  }%
}

\contourlength{0.8pt}

\newcommand{\fancyunderline}[1]{%
  \uline{\phantom{#1}}%
  \llap{\contour{white}{#1}}%
}

\begin{document}

\title{Phi-4 Technical Report}
\author{
\begingroup
\setlength{\tabcolsep}{10pt}
\begin{tabular}{cccc}
Marah Abdin & Jyoti Aneja & Harkirat Behl & S\'ebastien Bubeck \\
Ronen Eldan & Suriya Gunasekar & Michael Harrison & Russell J. Hewett \\
Mojan Javaheripi & Piero Kauffmann & James R. Lee & Yin Tat Lee \\
Yuanzhi Li & Weishung Liu & Caio C. T. Mendes & Anh Nguyen \\
Eric Price & Gustavo de Rosa & Olli Saarikivi & Adil Salim \\
Shital Shah & Xin Wang & Rachel Ward & Yue Wu \\
Dingli Yu & Cyril Zhang & Yi Zhang &
\end{tabular}
\endgroup
}
\date{Microsoft Research}
\maketitle

\begin{abstract}
We present \textbf{\modelwithoutspace}, a 14-billion parameter language model developed with a training recipe that is centrally focused on data quality. Unlike most language models, where pre-training is based primarily on organic data sources such as web content or code, \model strategically incorporates synthetic data throughout the training process. While previous models in the Phi family largely \emph{distill} the capabilities of a teacher model (specifically GPT-4), \model substantially \emph{surpasses} its teacher model on STEM-focused QA capabilities, giving evidence that our data-generation and post-training techniques go beyond distillation. Despite minimal changes to the \phithree{} architecture, \model achieves strong performance relative to its size -- especially on reasoning-focused benchmarks -- due to improved data, training curriculum, and innovations in the post-training scheme. 
\end{abstract}

\section{Introduction}

Recent advancements in Large Language Models (LLMs) have shown that significant improvements in data quality can rival, and sometimes surpass, the performance gains traditionally achieved by scaling compute with model and dataset size. Building on the success of the {Phi} family \cite{gunasekar2023textbooks,li2023textbooks,javaheripi2023phi,abdin2024phi}, we introduce \modelwithoutspace, a 14-billion parameter model that further advances performance of small language models by introducing innovative synthetic data generation methods for reasoning-focused tasks, by optimizing the training curriculum and data mixture, and by introducing new techniques in post-training.

Synthetic data constitutes the bulk of the training data for \model and is generated using a diverse array of techniques, including multi-agent prompting, self-revision workflows, and instruction reversal. These methods enable the construction of  datasets that induce stronger reasoning and problem-solving abilities in the model, addressing some of the weaknesses in traditional unsupervised datasets. Synthetic data in \model also plays a crucial role in post-training, where techniques such as rejection sampling and a novel approach to Direct Preference Optimization (DPO) are employed to refine the model’s outputs.

The development of \model is guided by three core pillars: \begin{enumerate} 
\item 
\textbf{Synthetic Data for Pretraining and Midtraining:} High-quality synthetic datasets are designed to prioritize \emph{reasoning} and \emph{problem-solving}, carefully generated to ensure diversity and relevance. We change our training curriculum and create new pretraining and midtraining data mixtures to increase the allocation of synthetic tokens, compared to older generations of \textsf{phi}.

\footnotetext[1]{These scores are lower than those reported by Meta, perhaps because \textsc{simple-evals} has a strict formatting requirement that Llama models have particular trouble following.  We use the \textsc{simple-evals} framework because it is reproducible, but Meta reports 77 for MATH and 88 for HumanEval on Llama-3.3.}
\stepcounter{footnote}

\item 
\textbf{Curation and Filtering of High-Quality Organic Data:} We meticulously curate and filter organic\footnote{We use \textit{organic} to refer to human-generated or otherwise non-synthetic data.} data sources, including web content, licensed books, and code repositories to extract seeds for the synthetic data pipeline that encourage high-depth reasoning and prioritize educational value (to the model). These seeds form the foundation of the synthetic generation pipeline. 
To complement these synthetic datasets, we also filter the web for high-quality data (in terms of knowledge and reasoning) to use directly in pretraining.

\item \textbf{Post-Training:} We further advance the post-training recipe in \model by creating new refined versions of SFT datasets, as well as by developing a new technique to create DPO pairs, based on \emph{pivotal token search}.

\end{enumerate}

With these innovations, the performance of \model on reasoning-related tasks is comparable to or surpasses much larger models. For example, its performance on many widely used reasoning-related benchmarks meets or exceeds that of \text{Llama-3.1-405B}. In Table \ref{tbl:benchmarks} we compare the performance of our model on academic benchmarks to several contemporary foundation models. We find that \model significantly exceeds its teacher GPT-4o on the GPQA (graduate-level STEM Q\&A) and MATH (math competition) benchmarks.

\renewcommand{\arraystretch}{1.2}
\begin{table}[t!]
\centering
\small
\begin{tabular}{@{}cc cccc ccc@{}}
\toprule
 & & \multicolumn{4}{c}{\textbf{Small models}} & \multicolumn{3}{c}{\textbf{Large models}}\\
\cmidrule(lr){3-6}\cmidrule(lr){7-9}
 & & \makecell{\textbf{phi-4}\\14b} & \makecell{\textbf{phi-3}\\14b} & \makecell{\textbf{Qwen 2.5}\\14b instruct} & \makecell{\textbf{GPT}\\4o-mini} & \makecell{\textbf{Llama-3.3}\\70b instruct} & \makecell{\textbf{Qwen 2.5}\\72b instruct} & \makecell{\textbf{GPT}\\4o}\\
\midrule
\multirow{7}{*}{\rotatebox[origin=c]{90}{\textbf{simple-evals}}} 
 & MMLU & 84.8 & 77.9 & 79.9 & 81.8 & 86.3 & 85.3 & \textbf{88.1}\\
 & GPQA & \textbf{56.1} & 31.2 & 42.9 & 40.9 & 49.1 & 49.0 & 50.6\\
 & MATH & \textbf{80.4} & 44.6 & 75.6 & 73.0 & 66.3\footnotemark[1] & 80.0 & 74.6\\
 & HumanEval & 82.6 & 67.8 & 72.1 & 86.2 & 78.9\footnotemark[1] & 80.4 & \textbf{90.6}\\
 & MGSM & 80.6 & 53.5 & 79.6 & 86.5 & 89.1 & 87.3 & \textbf{90.4}\\
 & SimpleQA & 3.0 & 7.6 & 5.4 & 9.9 & 20.9 & 10.2 & \textbf{39.4}\\
 & DROP & 75.5 & 68.3 & 85.5 & 79.3 & \textbf{90.2} & 76.7 & 80.9\\
\midrule
 & MMLUPro & 70.4 & 51.3 & 63.2 & 63.4 & 64.4 & 69.6 & \textbf{73.0}\\
 & HumanEval+ &82.8 & 69.2 &79.1 & 82.0 &77.9 & 78.4 & \textbf{88.0}\\
 & ArenaHard &75.4 &45.8 &70.2 & 76.2 &65.5 &\textbf{78.4} &75.6 \\
 & LiveBench &47.6 &28.1 &46.6 &48.1 &\textbf{57.6} & 55.3 &\textbf{57.6}\\
 & IFEval &63.0 &57.9 &78.7 &80.0 &\textbf{89.3} & 85.0 &84.8\\
\midrule
 & \makecell{PhiBench\\(internal)} &56.2 &43.9 &49.8 &58.7 &57.1 &64.6 & \textbf{72.4}\\
\bottomrule
\end{tabular}
\caption{Performance of \model{} on a set of standard benchmarks. The first set of benchmarks uses OpenAI's \textsc{simple-evals} framework~\cite{simple-evals}, specifying the prompts/extraction/temperature=0.5. We compare to small models of similar inference cost, as well as to larger models.
}
\label{tbl:benchmarks}
\end{table}

\subsection{Addressing Overfitting and Data Contamination}

\paragraph{Decontamination:}
One pitfall of foundation models is overfitting to benchmarks, such as through the leakage of benchmark test sets via the web corpus.  We improved the data decontamination process for \model compared to previous Phi models to ensure no unfair influence on evaluation results. More details of the decontamination method are given in Appendix \ref{sec:data_processing}.

\paragraph{AMC Benchmark:}  
The surest way to guard against overfitting to the test set is to test on fresh data. We tested our model on the November 2024 AMC-10 and AMC-12 math competitions~\cite{amc}, which occurred after all our training data was collected, and we only measured our performance after choosing all the hyperparameters in training our final model.  These contests are the entry points to the Math Olympiad track in the United States and over 150,000 students take the tests each year. 

\begin{figure}
    \centering
    \includegraphics[width=0.85\linewidth]{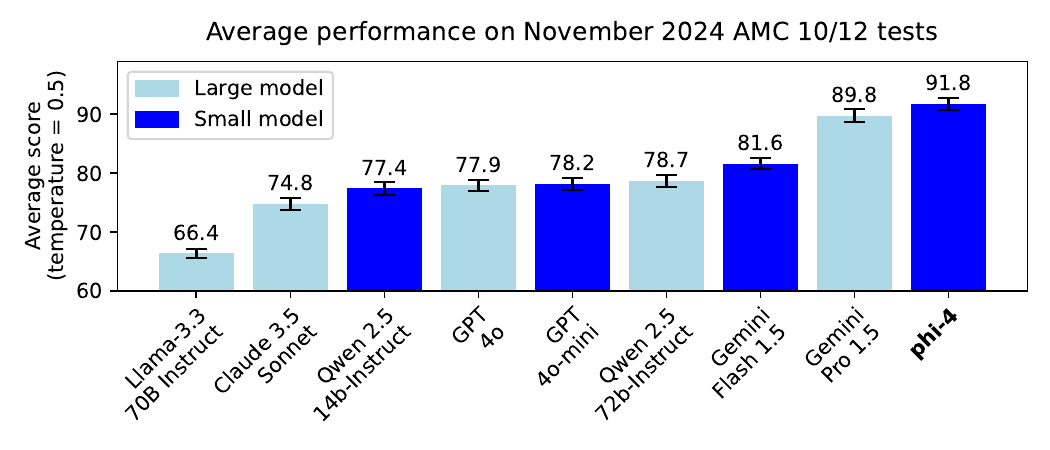}
    \caption{Average performance of different models on the November 2024 AMC-10 and AMC-12 tests.  This is the average score (with maximum score 150) over the four tests on 100 runs with temperature $t=0.5$. We chose $t = 0.5$ to follow \textsc{simple-evals}~\cite{simple-evals}.  Error bars are $2\sigma$ of the estimate.  On competition math, \model scores well above its weight-class even compared to non--open-weight models.}
    \label{fig:AMC}
\end{figure}

In Figure~\ref{fig:AMC} we plot the average score over the four versions of the test, all of which have a maximum score of 150.  \model outperforms not only similar-size or open-weight models but also much larger frontier models. Such strong performance on a fresh test set suggests that \modelwithoutspace's top-tier performance on the MATH benchmark is not due to overfitting or contamination. We provide further details in Appendix~\ref{sec:amc-appendix-details}.

\paragraph{Relying on Contamination-Proof Benchmarks:}
We give significant weight to benchmarks which were designed in such a way that the questions are original and do not appear on the web, such as GPQA~\cite{rein2023gpqa}. While optimizing our model, we relied on an internal benchmark composed primarily of original prompts written by the team (see Section \ref{sec:phibench} for further details).

\paragraph{Long Chain-of-Thought Models:}
A style of LLM that scales inference-time compute by generating long chains of thought has emerged over the past few months, as pioneered by OpenAI O1~\cite{openai_learning_to_reason} and followed by DeepSeek-R1-Lite-Preview~\cite{deepseek_r1_lite_preview} and Qwen/QwQ-32B-Preview~\cite{qwq-32b-preview}.  These models perform well on reasoning benchmarks, where QwQ, the only such model with open weights, averages 124.5 points in the AMC-10/12 setting of Figure~\ref{fig:AMC}.  However, QwQ also uses 4X more tokens on this task than \model and has more than twice as many parameters.  Thus, the inference cost of QwQ is an order of magnitude higher than \modelwithoutspace.  Consequently, these models are not in the same class as \model with respect to cost or latency.

\section{Approach to Data}
The pretraining phase of \model relies heavily on synthetic datasets generated through a variety of techniques. In addition, we employ several methods for filtering organic data sources that are used both as complementary datasets in the pretraining and as seeds for generating synthetic data.

\subsection{Purpose of Synthetic Data}

Synthetic data as a substantial component of pretraining is becoming increasingly common, and the Phi series of models has consistently emphasized the importance of synthetic data.  Rather than serving as a cheap substitute for organic data, synthetic data has several direct advantages over organic data.

\paragraph{Structured and Gradual Learning.} In organic datasets, the relationship between tokens is often complex and indirect. Many reasoning steps may be required to connect the current token to the next, making it challenging for the model to learn effectively from next-token prediction. By contrast, each token generated by a language model is by definition predicted by the preceding tokens, making it easier for a model to follow the resulting reasoning patterns. In this way, synthetic data may act as a form of ``spoonfeeding," presenting challenges in a digestible and progression-oriented manner.

A simple example to illustrate this is that a human-written solution to a math problem might start with the final answer.  This answer is much too hard to output immediately, for either a human or an LLM---the human produced it by nonlinear editing, but pretraining expects the LLM to learn to produce it linearly.  Synthetic solutions to math problems will not have such roadblocks.

\paragraph{Alignment with Inference Contexts.} Synthetic data is typically closer to the format of outputs we expect our models to generate. Training on such data helps align the model’s pretraining experience with the scenarios it encounters during inference. This alignment ensures that the context seen during generation remains in-distribution with respect to the data the model was pretrained on.

For example, web forums are very different in style from LLM interactions.  If a fact only appears in web forum data, the pretrained model will think it is very unlikely to occur in the chats it produces.  Rewriting facts from the web forum into the language style of an LLM makes the facts more accessible during the LLM chat context of inference.

\paragraph{Principles.} Our approach to generating synthetic data for \model is guided by the following principles:
\begin{enumerate}
    \item \textbf{Diversity:} The data should comprehensively cover subtopics and skills within each domain. This requires curating diverse seeds from organic sources.
    \item \textbf{Nuance and Complexity:} Effective training requires nuanced, non-trivial examples that reflect the complexity and the richness of the domain. Data must go beyond basics to include edge cases and advanced examples.
    \item \textbf{Accuracy:} Code should execute correctly, proofs should be valid, and explanations should adhere to established knowledge, etc.
    \item \textbf{Chain-of-Thought:} Data should encourage systematic reasoning, teaching the model various approaches to the problems in a step-by-step manner. This fosters coherent outputs for complex tasks.
\end{enumerate}

\subsection{Synthetic Data for Pretraining and Midtraining}

We created 50 broad types of synthetic datasets, each one relying on a different set of seeds and different multi-stage prompting procedure, spanning an array of topics, skills, and natures of interaction, accumulating to a total of about 400B unweighted tokens. In Appendix \ref{app:synthexamples}, we give a few examples of transcripts taken from our synthetic generations.
Here, we highlight novel methodologies used in generating synthetic datasets for \modelwithoutspace:
\begin{itemize}
\item \textbf{Seed Curation:} The synthetic dataset generation begins with high-quality seeds sourced from multiple domains.  These curated seeds provide the foundation for synthetic data generation, enabling the creation of exercises, discussions, and reasoning tasks tailored to the model's training objectives.
\begin{enumerate}
    \item \textbf{Web and Code-based Seeds:} Excerpts and snippets are extracted from web pages, books, and code repositories with a focus on content that demonstrates high complexity, reasoning depth, and educational value. To ensure quality, we employ a two-stage filtering process: first, identifying pages with strong educational potential, and second, segmenting the selected pages into passages, scoring each for its factual and reasoning content.
    \item \textbf{Question Datasets:} A large set of questions was collected from websites, forums, and Q\&A platforms. These questions were then filtered using a plurality-based technique to balance difficulty. Specifically, we generated multiple independent answers for each question and applied majority voting to assess the consistency of responses. We discarded questions where all answers agreed (indicating the question was too easy) or where answers were entirely inconsistent (indicating the question was too difficult or ambiguous). This filtering process produces a dataset of questions that challenge the model's reasoning and problem-solving abilities while remaining approachable. The plurality answers were used in place of the ground truth in our rejection-sampling based generations. 
    \item \textbf{Creating Question-Answer pairs from Diverse Sources:} Another technique we use for seed curation involves leveraging language models to extract question-answer pairs from organic sources such as books, scientific papers, and code. This approach does not rely on merely identifying explicit Q\&A pairs within the text. Instead, it involves a pipeline designed to detect deduction chains or logical progressions in the text. The language model identifies key steps in reasoning or problem-solving processes and reformulates them into questions and corresponding answers. Our experiments show that, if done correctly, training on the resulting content can be far more effective (in terms of improvement on academic and internal benchmarks) than training on the original content. 
\end{enumerate}
    \item \textbf{Rewrite and Augment:} Seeds are transformed into synthetic data through multi-step prompting workflows. This includes rewriting most of the useful content in given passages into exercises, discussions, or structured reasoning tasks. 
    \item \textbf{Self-revision:} The initial responses are then iteratively refined through a feedback loop where a model critiques and subsequently improves its own outputs, guided by the rubrics focused on reasoning and factual accuracy.
    \item \textbf{Instruction Reversal for Code and Other Tasks:} To enhance the model’s ability to generate outputs from instructions, we used an instruction reversal technique. For example, we take existing code snippets from the code data corpus and use it to generate corresponding instructions that include the problem description or task prompt. The resulting synthetic data pairs were structured with the instruction appearing before the code. Only data with high fidelity between the original and regenerated code are retained, ensuring alignment between the instructions and the outputs. This method can be generalized to other targeted use cases. 
    \item \textbf{Validation of Code and Other Scientific Data:} When appropriate, we incorporate tests for validating our reasoning-heavy synthetic datasets. The synthetic code data is validated through execution loops and tests. For scientific datasets, the questions are extracted from scientific materials using a method designed to ensure high relevance, groundedness, and difficulty balance. 
\end{itemize}

\subsection{Curation and Filtering of Web and Q\&A Data}
\paragraph{Q\&A datasets.}
We collected tens-of-millions high-quality organic problems and solutions by reviewing public websites, relying on existing datasets, and acquiring external datasets. Our experience from previous models showed that question-answer data contributed significantly to various capabilities, such as mathematical reasoning and academic performance. Our ablation studies showed that organic questions are substantially more effective than synthetic questions. We used several ways to synthetically augment the dataset of organic questions to obtain a larger dataset. While these rewritten questions improved the model’s capabilities, the gains were not as pronounced.
A significant portion of the collected questions lacked accurate solutions. To address this, we replaced the answers with synthetically generated ones and used majority-voting to increase accuracy.
All collected questions and solutions underwent a thorough decontamination process to ensure there is no overlap with test sets\footnote{This step is crucial to the reliability of some of the academic benchmarks: for instance, some test benchmark variants can be found on platforms like Hugging Face. Moreover, benchmarks such as MMLU are frequently compiled from web-sourced questions.}.

\paragraph{Targeting High-quality Web Data.}\label{sec:web_data}
We collected a wide variety of high-quality organic data sources for \modelwithoutspace, prioritizing reasoning-dense and nuanced material (e.g., academic papers, educational forums, and programming tutorials). In addition to directly training on this text, we used various web sources as seeds for specialized synthetic data generation pipelines. We found clean and correct natural data to be absolutely crucial for seeding synthetic data: minor errors can result in severe quality degradations for derived synthetic documents. We therefore invested heavily in the \textit{perfectionistic} curation of our web data. We discuss the main techniques and considerations below:

\begin{itemize}
\item \textbf{Targeted Acquisitions:} We included major repositories of reasoning-dense documents that are publicly permissible for use (e.g., arXiv, PubMed Central, GitHub) or explicitly licensed (e.g., licensed books) aiming for a level of comprehensiveness, recency, and cleanliness above the typical standard of externally available corpora. 
\item \textbf{Filtering Web Dumps:} To capture the long tail of information-rich web sources (e.g., forums, blogs, course material, domain-specific wikis), we took the approach of selecting a small fraction of highest-quality documents from bulk web dumps, using small (non-LLM) classifiers trained on $\sim10^6$ LLM-generated annotations. This approach tends to over-index on STEM-related keywords, so we created a specialized pipeline to amplify high-quality non-STEM content (e.g., arts, history, travel, culture, and entertainment). These topic classifications were also obtained by distilling an LLM annotator. Finally, we removed corrupted text and binary files by detecting outliers according to $n$-gram statistics and compression ratios.
\item \textbf{Multilingual Data:} We incorporated multilingual datasets to ensure that our model could handle a wide range of languages, including German, Spanish, French, Portuguese, Italian, Hindi and Japanese. This involved sourcing and processing high-quality multilingual documents from CommonCrawl and Wikipedia. Our multilingual processing pipeline consists of a language identification model, based on \texttt{fastText} used to categorize documents into 176 languages, then uses the same classifiers for filtering web dumps to filter for quality. Note that the classifiers were trained on multilingual LLM-generated annotations.

\item \textbf{Custom Extraction and Cleaning Pipelines:} To ensure sufficient cleanliness and uniformity between heterogeneous organic data sources, we needed a collection of customized heuristics and parsers. For each targeted data source, we built custom pipelines to ingest a variety of file formats (e.g., multi-file TeX source, ePub and other XML-like formats, Microsoft Word documents, and PDFs). For general web data, we built a custom HTML-to-text extractor, taking significant care to preserve fragile content that is frequently corrupted by na\"ive parsers (e.g., TeX/MathML equations, code blocks, tables, and forum thread structure). This extractor prunes and normalizes the DOM tree, using a variety of signals (e.g., HTML tag names, CSS classes, content length, and tree depth) to distinguish elements such as boilerplate, advertisements, equations, and syntax-highlighter artifacts.
\end{itemize}

\subsection{Post-Training datasets} 
Our post-training data is composed of:
\begin{itemize}
    \item \textbf{Supervised Fine-Tuning (SFT) Datasets:} Using carefully curated user prompts taken from a mixture of publicly available datasets and synthetically generated data, we generate multiple model responses and select the best using an LLM-based evaluation process. 
    \item \textbf{Direct Preference Optimization (DPO):} We generate DPO pairs based on rejection sampling and LLM evaluation, a part of which is based on our approach to creating pivotal token-based pairs, explained in Section~\ref{sec:pivotal} below.
\end{itemize}

\section{Pretraining details}\label{sec:pretrain}
The \model model is based on a decoder-only transformer architecture \cite{Vas17} with $14$B parameters and a default context length of $4096$.  This is later extended to a 16K context length during midtraining. The architecture closely follows \phithree{}-medium, except that we now use the \texttt{tiktoken} tokenizer (for better multilingual support) with a padded vocabulary size of {100,352} (including unused tokens) and we use full attention over the $4$K context length, rather than a $2$K sliding window  used in \phithreemed.

The model was pretrained for approximately $10$T tokens using linear warm-up and decay schedules with peak learning rate of $0.0003$, constant weight decay of $0.1$, and global batch size of $5760$. The training hyperparameters are tuned using interpolations from shorter horizon runs and further adjusted by stress testing the learning rate warm-up stage for stability. Pretraining is followed by a shorter midtraining stage to increase the original context length of $4$k to $16$k.

Since pre-trained models are not good at instruction following, it is not very informative to use 0-shot evaluations that require the answer to be in a specific format, for example \textsc{simple-evals}. We therefore use an internal implementation of benchmarks for pretraining which uses a mixture of log-likelihood and/or few-shot prompts for various tasks. Specifically, we used log-likelihood evaluations for MMLU (5-shot), MMLU-pro, and ARCC (1-shot). We used 1, 3, 4, and 8 few-shot examples for TriviaQA (TQA), MBPP, MATH, and GSM8k to help the model adhere to the answer format for easier extraction of the solution. We use this evaluation method throughout Section~\ref{sec:pretrain}. Table ~\ref{tab:phi3_versus_phi4} summarizes the performance boost of pretrained \model compared with its predecessor \phithree{}-medium.

\begin{table}[]
    \centering
\resizebox{0.9\linewidth}{!}{
\begin{tabular}{c c c c c c c c c}
\toprule
 & MMLU & MMLU pro & GSM8k & Human-Eval & ARCC & MBPP & MATH & TQA \\ \hline
\model (4k) & +3.0 &	+10.3&		+2.2&		+7.8	&	+1.1	&	+6.8	&	+8.9 &		-0.7\\
\model (16k) & +2.7	&+8.9	&+1.2	&+9.0	&+0.9	&+9.6	&+8.4	&-1.5\\
\bottomrule
\end{tabular}}
    \caption{Pretraining benchmarks for \model compared to its predecessor, \phithree{}-medium after pretraining.}
    \label{tab:phi3_versus_phi4}
\end{table}

\subsection{Data Composition in Pretraining}\label{sec:data_comp}

The \phithree{} model family were trained using a two-phase strategy. Most of the training tokens were used in phase 1 of the training, which consisted largely of filtered web data. Phase 2 was trained with a data mixture consisting primarily of synthetic tokens and a much smaller allocation for ultra-filtered and reasoning-heavy web data. As the size and complexity of our synthetic data grew, we observed a marginal drop in the benefit from using non-synthetic tokens for the \phithree{} family of model sizes. We note two key observations.
\begin{itemize}
    \item Web datasets showed small benefits on reasoning heavy benchmarks. Prioritizing more epochs over our synthetic data led to better performance with respect to adding fresh web tokens. 
    \item Models trained only with synthetic data underperformed on the knowledge-heavy benchmarks and demonstrated increased hallucinations. 
\end{itemize}

\begin{figure}[ht]
    \centering
    \includegraphics[width=0.6\linewidth]{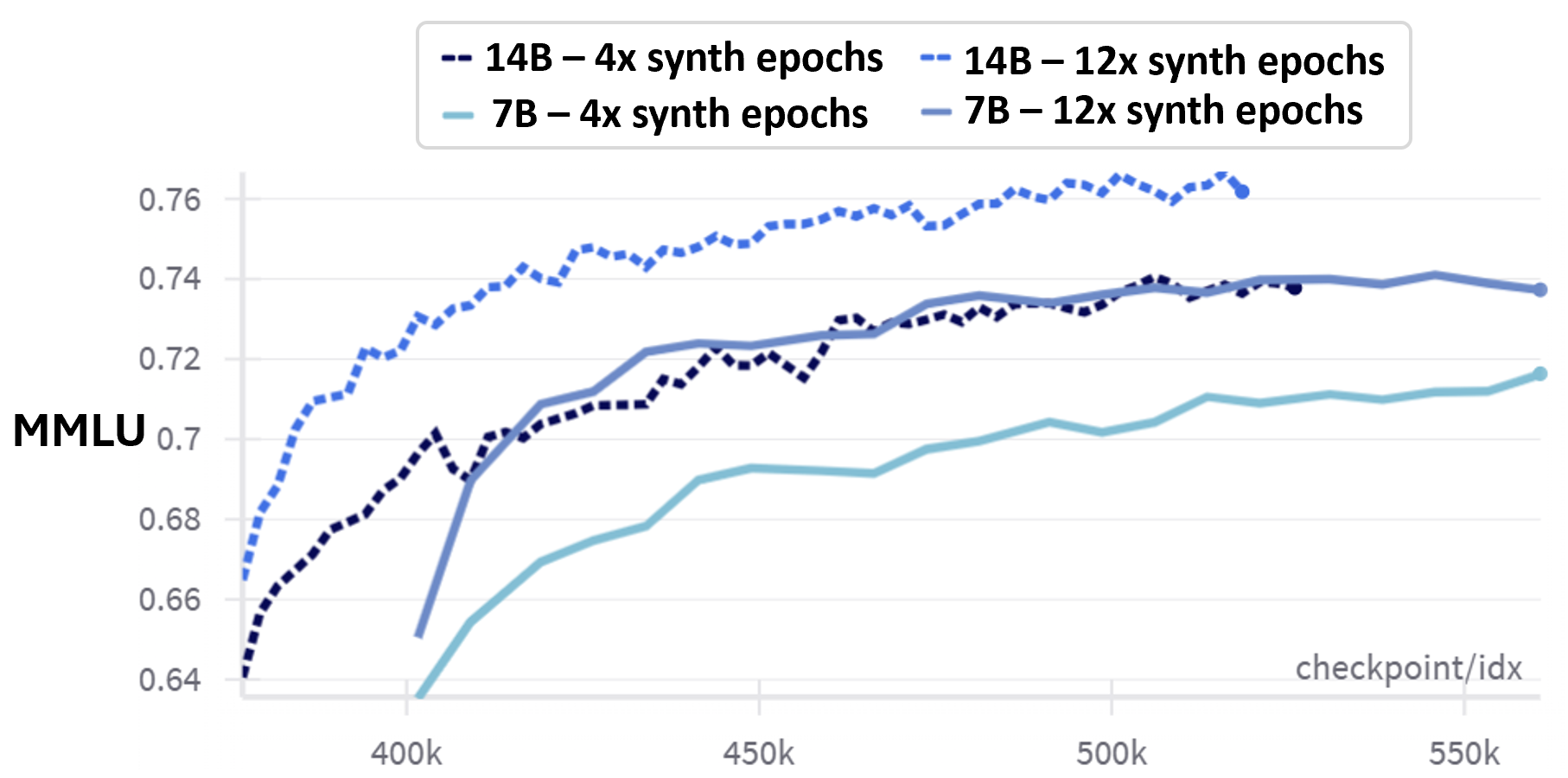}
    \caption{5-shot MMLU score for phase 2 pretraining runs with 4 and 12 epochs of synthetic data. All models are trained for the same token horizon, thus the model with 4 epochs of synthetic has seen more (unique) web tokens. We see that despite many epochs on synthetic data, we do not see overfitting behavior and in fact the 12 epoch models perform better than those that have seen more unique web tokens.}
    \label{fig:synth_epochs}
\end{figure}

Figure~\ref{fig:synth_epochs} demonstrates the first phenomenon using smaller scale phase 2 pretraining exercises. In this example, we conduct two training runs per model scale, using the same number of training tokens on top of phase 1 pretrained checkpoints. For all runs, the number of unique synthetic tokens is fixed (a subsample of full synthetic data) but the number of repetitions on this data changes, namely 4 and 12 epochs. The rest of the training tokens are fresh unique tokens supplied from web sources. As seen, performing more iterations on the synthetic data is more beneficial than supplying more web tokens.

Inspired by this scaling behavior of our synthetic data, we trained a $13$B parameter model solely on synthetic\footnote{This is an updated mixture of synthetic data that contains new sources compared to \phithree{}.} data, for ablation purposes only -- the model sees over $20$ repetitions of each data source. For the sake of ablations, we partitioned our synthetic data into \emph{web rewrites}, which includes more direct rewrites of our filtered web content relative to all other types of synthetic data.
Table~\ref{tab:synth_versus_phi3} compares the previous \phithreemed~model with the new model trained entirely on the synthetic data. Throughout training, all benchmarks consistently improved, despite the increase in epochs, and the majority of the benchmarks showed improvements over \phithree{}. However, knowledge-related benchmarks, like 1-shot triviaqa (TQA), show a large gap where synthetic models are subpar. These observations led us to rethink the role of web data in our data mixture.

\begin{table}[ht]
\centering
\resizebox{\linewidth}{!}{
\begin{tabular}{c c c c c c c c c}
\toprule
 & MMLU & MMLU pro & GSM8k & Human-Eval & ARCC & MBPP & MATH & TQA \\ 
\midrule
Synthetic & +0.8 & +4.0 & +2.2 & +12.1 & 0.0 & +5.0 & +4.9 & -14.8 \\
Synthetic + Web Rewrites & +0.3 & +4.1 & +1.8 & +13.3 & +3.0 & +7.6 & +8.1 & -7.7 \\
\bottomrule
\end{tabular}}
\caption{Benchmark performance of $13$B models (used for ablations only) trained on data mixtures containing no web data. The respective training tokens are either from synthetic sources, or an equal share of synthetic data and web rewrites. All numbers are reported relative to the performance of \phithreemed, which has seen a combination of web and synthetic data.}
\label{tab:synth_versus_phi3}
\end{table}

\subsection{Data Mixture}
To design our pretraining data mixture for a given training token budget, we search over different allocation of tokens coming from various sources, namely, 1) synthetic, 2) web rewrites\footnote{Web rewrites is a sub-category of synthetic data that is substantially large and contains direct rewrites of web content.}, 3) filtered web (divided into reasoning and knowledge-heavy portions), 4) targeted acquisitions and organic data (e.g., academic data, books, and forums), and 5) code data. 

We conducted ablations using a shorter token horizon of $1$T tokens to derive the data mixture. These ablations rely on our established result on the high-rank correlation of short training with longer training, up to the over-fitting saturation threshold of data sources. In addition we observe a high rank correlation between the performance of the $7$B and $14$B models on different data mixtures, given a large enough distance between the data mixtures. This allowed us to conduct the experiments at $7$B scale and transfer the findings to \modelwithoutspace. Among the numerous ablations, we highlight a few that show best insights on our data composition. Specifically, we freeze the ratio of tokens coming from targeted acquisitions and code categories, and change the ratio of tokens for the synthetic, web, and web rewrites clusters.

\begin{table}[ht]
\centering
\resizebox{0.95\linewidth}{!}{
\begin{tabular}{c c c c c c c c c c}
\toprule
 & MMLU & MATH & GSM8k & Human-Eval & ARCC & MBPP & TQA & MMLU pro & Average \\
\midrule
Uniform & -3.3 & -5.4 & -5.8 & -1.2 & +0.6 & -2.0 & +3.3 & -3.6 & -2.2 \\
S & +3.3 & +4.0 & +2.1 & -6.1 & +1.9 & +0.4 & -3.0 & +3.7 & +0.8 \\
S + WR & +0.6 & +1.2 & +1.5 & -1.2 & +1.6 & +1.6 & -3.7 & +1.2 & +0.4 \\
S + W & -0.6 & -0.7 & -0.7 & -4.3 & +0.3 & -2.0 & +6.9 & +0.9 & 0.0 \\
\bottomrule
\end{tabular}}
\caption{Ablations on the allocation of $75\%$ of training tokens to synthetic (S), filtered web (W), and web rewrite (WR) categories, while other data sources are held constant in the remaining $25\%$ token budget. All benchmark numbers are measured relative to the final data mixture used for training \modelwithoutspace.}
\label{tab:data_mixture_ablation}
\end{table}

Table~\ref{tab:data_mixture_ablation} summarizes the results for the hand-picked ablations, as compared with the data mixture that was used for the final training run. A uniform allocation of tokens among the three categories is suboptimal due to the higher quality of synthetic data and the only benchmark that shows a clear benefit from web data is TQA. While the synthetic-heavy variations on rows 2 and 3 of the table are marginally better than the chosen final data mixture, we decided to integrate the targeted and knowledge-heavy filtered web data sources to improve knowledge benchmarks (see Section~\ref{sec:data_comp}) to balance all model capabilities. We also note that we observed the gap between the chosen data mixture and the synthetic heavy runs largely closes as the model goes through the post-training stage. An end-to-end optimization of pretraining data mixture that also takes into account the effects of post-training is an interesting future area of investigation.

\begin{table}[ht]
\centering
\resizebox{0.55\linewidth}{!}{
\begin{tabular}{c c c c}
\toprule
\makecell{Data\\Source} & \makecell{Fraction\\of Training} & \makecell{Unique\\Token Count} & \makecell{Number of\\Epochs}\\
\midrule
Web & 15\% & 1.3T & 1.2 \\
Web rewrites & 15\% & 290B & 5.2 \\
Synthetic & 40\% & 290B & 13.8 \\
Code data & 20\% & 820B & 2.4 \\
Acquired sources & 10\% & 580B & 1.7 \\
\bottomrule
\end{tabular}}
\caption{Data mixture for pretraining.}
\end{table}

The final data mixture used for \model allocates $30\%$ of the training tokens to web and web rewrites data sources, divided equally between them. The remaining tokens are largely sourced from synthetic data which accounts for $40\%$ of the data mixture tokens. Finally we allocate $20\%$ of tokens to code data (mixture of synthetic and raw code) and $10\%$ to targeted acquired sources like academic data and books. In terms of total number of unique tokens in each data mixture cluster, filtered web data is the largest cluster with $\sim1.3$T tokens. Code and targeted acquisitions are the second and third largest clusters with $\sim820$B and $\sim580$B tokens, respectively. Finally, web rewrites and synthetic data have similar token count of $\sim290$B tokens. The total number of epochs on each data source is determined using the ratio of allocated tokens in the mixture and the number of unique tokens in that source.

\subsection{Midtraining Details}

\model includes a midtraining stage where the context length is increased from the original $4$K to $16$K. We conduct several ablations to study the role of data on long-context performance. Specifically, we try data sources that are inherently long context, and compare them with artificially created long context data where samples are padded together to fill the sequence. We observe the former to perform better in longer context tasks. 

Inspired by this, we further filter our high-quality non-synthetic datasets (i.e., academic, books, and code data) to separate samples above $8$K context. We then up-weight the data subsets that are $16$K or higher in length. We also create new synthetic datasets that satisfy the $>4$K sequence requirement. The final data mixture includes $30\%$ of the newly curated longer context data and a $70\%$ portion of recall tokens from the pretraining stage. To accommodate longer context, we increase the base frequency of rope position encoding to $250$K following~\cite{llama3report}. We drop the maximum learning rate by a factor of $10$ compared to the pretraining stage and train for a total of $250$B tokens.

To effectively evaluate the long-context capability of our model, it is essential to have a comprehensive evaluation framework with practical scenarios. While synthetic benchmarks like needle-in-a-haystack and RULER are preferred for their simplicity and control, our emphasis is on a diverse range of tasks that reflect real-world applications, such as reasoning across entire documents. 
We report the performance of \model and other models on the tasks we selected from the HELMET~\cite{yen2024helmetevaluatelongcontextlanguage} evaluation suite in Table~\ref{tbl:helmet} and outline our evaluation methods below. Note that results are average across 5 runs for each categories.

\begin{itemize}
    \item Recall: The task involves retrieving the corresponding value from a randomly-generated long JSON file given a specific key (Metric: SubEM)
    \item RAG: Answer questions based on many retrieved and shuffled Wikipedia documents. The datasets used for this task are NaturalQuestions, HotpotQA, and PopQA. Final results are average of all datasets (Metric: SubEM)
    \item Re-rank: The task is to re-rank the top-10 documents given a query and many retrieved and shuffled documents. This uses the MSMARCO dataset (Metric: nDCG@10)
    \item ICL: The task involves many-shot in-context learning with datasets such as TREC coarse, TREC fine, Banking77, NLU and CLINC150. Final results are average of all datasets (Metric: F1)
    \item QA: Answer questions given a lengthy document. The dataset associated with this task is NarrativeQAv2 (Metric: GPT-4o scoring)
    \item Summ: The task involves summarizing a lengthy legal document, and the dataset used is MultiLexSum (Metric: GPT-4o scoring)
\end{itemize}

\begin{table}[t!]
\centering
\small
\begin{tabular}{@{}cccccccc@{}}
\toprule
\textbf{Model} & \textbf{Max Length} & \textbf{Recall} & \textbf{RAG} & \textbf{ICL} & \textbf{Re-rank} & \textbf{QA} & \textbf{Summ} \\
\midrule
phi-4 & 8K & 100.0 & 58.1 & 68.0 & 65.3 & 26.7 & 38.3 \\
Qwen-2.5-14B & 8K & 100.0 & 62.2 & 67.8 & 58.2 & 24.7 & 37.2 \\
Llama-3.3-70B & 8K & 92.0 & 65.3 & 69.4 & 64.4 & 30.0 & 37.8 \\
GPT-4o-mini & 8K & 99.2 & 65.8 & 74.4 & 69.4 & 31.3 & 38.5 \\
GPT-4o & 8K & 100.0 & 66.9 & 83.0 & 75.1 & 37.3 & 43.0 \\
\midrule
phi-4 & 16K & 99.0 & 57.1 & 77.0 & 54.4 & 36.0 & 40.5 \\
Qwen-2.5-14B & 16K & 100.0 & 59.1 & 67.6 & 50.3 & 29.7 & 42.3 \\
Llama-3.3-70B & 16K & 92.0 & 62.2 & 70.0 & 63.3 & 36.7 & 41.9 \\
GPT-4o-mini & 16K & 100.0 & 63.6 & 78.4 & 63.9 & 36.0 & 45.2 \\
GPT-4o & 16K & 100.0 & 66.7 & 85.6 & 73.8 & 43.7 & 46.3 \\
\bottomrule
\end{tabular}
\caption{Evaluation results on the long-context benchmark HELMET~\cite{yen2024helmetevaluatelongcontextlanguage}.}
\label{tbl:helmet}
\end{table}

\section{Post-Training}
Post-training is aimed at transforming the pretrained language model into an AI assistant that users can safely interact with. 
We align the pretrained model with one round of SFT \ref{sec:sft}, one round of DPO~\cite{rafailov2023direct} on data from our pivotal token search method (see Section~\ref{sec:pivotal}), and one round of DPO on full length preference pairs.
The model is chat finetuned using the standard \texttt{chatml} format, example usage template for two rounds of a conversation is as follows:
\begin{AIbox}{}
\tt \footnotesize 
<\!|\!im\_start\!|\!>system<\!|\!im\_sep\!|\!>system\_message<\!|\!im\_end\!|\!>\\
<\!|\!im\_start\!|\!>user<\!|\!im\_sep\!|\!>prompt1<\!|\!im\_end\!|\!><\!|\!im\_start\!|\!>assistant<\!|\!im\_sep\!|\!>response1<\!|\!im\_end\!|\!>\\
<\!|\!im\_start\!|\!>user<\!|\!im\_sep\!|\!>prompt2<\!|\!im\_end\!|\!><\!|\!im\_start\!|\!>assistant<\!|\!im\_sep\!|\!>
\end{AIbox}

\subsection{Supervised Fine-Tuning} \label{sec:sft}
In this phase, we fine-tune the pretrained model with a learning rate of $10^{-6}$ on a variety of data generated from high-quality data across diverse domains, including math, coding, reasoning, conversation, model identity, and safety.
We also added multilingual data for 40 languages.
We use around 8B tokens of data in this phase, all formatted in the \texttt{chatml} format.

\subsection{Direct Preference Optimization}
We use DPO \cite{rafailov2023direct} to align the model with human preferences, and also to steer the model away from unwanted behavior through pairs of desired and undesired outputs. 
DPO data covers chat format data, reasoning, and Responsible AI (RAI) data and improves the model in math, coding, reasoning, robustness, and safety.
We do two rounds of DPO on the SFT model. We introduce a technique, Pivotal Token Search (PTS), to generate pairs for DPO for the first DPO round. Details of the data mixture for first round are provided in Table \ref{tab:training_data_table_1}.

For the second round, which we call judge-guided DPO, we gather approximately 850k pairs of desired and undesired outputs. The prompts are sourced from various publicly available instruction tuning datasets and also include prompts related to safety and Responsible AI (RAI).
Next, for each of these prompts, we generate responses from GPT-4o, GPT-4t and our model. From these responses, we create various combinations of DPO pairs and use GPT-4o as a judge to label positive or negative for a given pair. For a given pair of responses, each assistant response is given a score based on accuracy, style, and detail. 
We label the response with higher accuracy or overall (average of accuracy, style, and detail) score as the positive response. We provide the prompt we used in Appendix~\ref{sec:post_train_appendix}. The data mixture for this round is provided in Table \ref{tab:training_data_table_2}.
Both stages also include a small amount of data for safety and mitigating hallucinations.

\begin{table}[t!]
\centering
\small
\begin{minipage}{0.48\textwidth}
\centering
\begin{tabular}{@{}cc@{}}
\toprule
\textbf{Dataset Name} & \textbf{Sample Count} \\
\midrule
unknown + safety data & 3,000 \\
generic multiple-choice Q\&A & 132,859 \\
math data & 76,552 \\
python data & 16,080 \\
cpp, go, java, js, rust data & 21,806 \\
\bottomrule
\end{tabular}
\caption{Data Mixture for Pivotal Token DPO}
\label{tab:training_data_table_1}
\end{minipage}
\hfill
\begin{minipage}{0.48\textwidth}
\centering
\begin{tabular}{@{}cc@{}}
\toprule
\textbf{Dataset Name} & \textbf{Sample Count} \\
\midrule
unknown + safety data & 43,842 \\
any vs any overall & 266,000 \\
any vs any accuracy & 532,000 \\
\bottomrule
\end{tabular}
\caption{Data Mixture for Judge Guided DPO}
\label{tab:training_data_table_2}
\end{minipage}
\end{table}

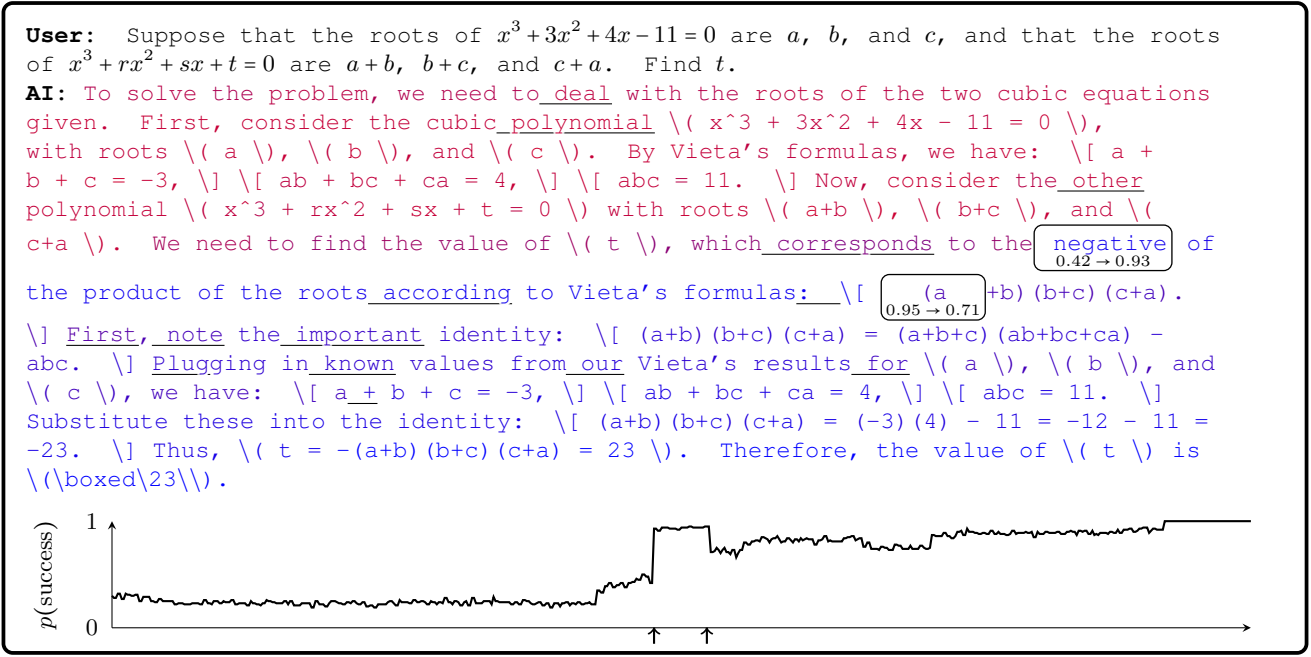
\begin{figure}
\begin{AIbox}{}
    \tt \footnotesize
    \begin{minipage}{0.95\textwidth}
    \include{pivotal_token_text}
    \vspace{-1em}
    \end{minipage}

\begin{tikzpicture}
    \begin{axis}[
        axis lines=left,
        ytick={0, 1},
        ytick style={draw=none},
        ylabel={$p(\mathrm{success})$},
        ylabel style={at={(0,0.5)}, anchor=center, yshift=-10pt},
        xtick=\empty,
        ymin=0, ymax=1,
        height=3cm,
        width=\textwidth,
        clip=false,
        enlargelimits=false,
    ]
    \addplot[
        color=black,
        thick,
    ]
    table[y index=0, x expr=\coordindex] {pivotal_token_probs.csv};

    \draw[->, thick] (axis cs:277, 0) -- ++(0, 0.1);
    \draw[thick, -] (axis cs:277, 0) -- (axis cs:277, -0.15);

    \draw[->, thick] (axis cs:304, 0) -- ++(0, 0.1);
    \draw[thick, -] (axis cs:304, 0) -- (axis cs:304, -0.15);
    \end{axis}
\end{tikzpicture}
    \vspace{-2.5em}
    \end{AIbox}
    \caption{Illustration of pivotal tokens for GPT-4o at temperature $1$ on a problem from the MATH benchmark~\cite{hendrycksmath2021}, where the
 initial success probability is $0.31$. Each token is colorized by the probability of success for an independent completion ($N=529$) continued from after the token, with \textcolor{red!100!black}{red for $p(\mathrm{success})=0$} and \textcolor{blue!100!black}{blue for $p(\mathrm{success})=1$}. The line plot shows the same probabilities. The tokens that changes $p(\mathrm{success})$ by $\ge 0.2$ are shown \tokenhilightbox{boxed}, with subscripts showing the change in probability. Tokens with probability $\leq 0.1$ are \underline{underlined} to illustrate that pivotal tokens are distinct from low-probability tokens. The token probabilities of \tokenhilightbox{~negative} and \tokenhilightbox{(a} were 0.31 and 0.12, respectively. The greedy tokens for the same prefixes are \tokenhilightbox{~product} with 0.66 probability and \tokenhilightbox{t} with 0.88 probability.}
    \label{fig:pivotal-token-example}
\end{figure}
\algnewcommand{\Yield}{\textbf{yield}}
\begin{figure}
    \begin{algorithmic}
        \Procedure{PivotalTokenSearch}{$Q,T_\mathrm{full},p_\mathrm{gap}$}
            \Procedure{Subdivide}{$T_\mathrm{prefix},T$}
                \If{$\left| T \right| \leq 1$ or $\left|\, p(\mathrm{success} \mid T_\mathrm{prefix}) - p(\mathrm{success} \mid T_\mathrm{prefix} + T) \,\right| < p_\mathrm{gap}$} \Comment{Base cases.}
                    \State \Return $[T]$
                \EndIf
                \State $T_\mathrm{left},T_\mathrm{right} \gets \Call{Split}{T}$ \Comment{We split at the cumulative midpoint of token log probabilities.}
                \State \Return $\Call{Subdivide}{T_\mathrm{prefix}, T_\mathrm{left}} \cup \Call{Subdivide}{T_\mathrm{prefix} + T_\mathrm{left}, T_\mathrm{right}}$
            \EndProcedure
            \State $T_\mathrm{prefix} \gets \epsilon$
            \ForAll{$T \in \Call{Subdivide}{\epsilon, T_\mathrm{full}}$}
                \If{$\left| T \right| = 1$ and $\left|\, p(\mathrm{success} \mid T_\mathrm{prefix}) - p(\mathrm{success} \mid T_\mathrm{prefix} + T) \,\right| \ge p_\mathrm{gap}$}
                    \State \Yield~$(Q,T_\mathrm{prefix},T)$ \Comment{Output pivotal tokens $T$ and context for postprocessing.}
                \EndIf
                \State $T_\mathrm{prefix} \gets T_\mathrm{prefix} + T$
            \EndFor
        \EndProcedure
    \end{algorithmic}
    \caption{Pseudocode for Pivotal Token Search (PTS). Note that estimating $p(\mathrm{success} \mid \dots)$ involves sampling the language model and invoking the oracle. In an efficient implementation $p(\mathrm{success} \mid \dots)$ should be memoized.}
    \label{fig:pts-pseudocode}
\end{figure}

\subsection{Pivotal Token Search} \label{sec:pivotal}

Consider a generative model producing a token-by-token response to a given prompt. For each token produced, which corresponds to a prefix of the model response, one can consider the conditional probability of the model's answer being correct given that prefix, as well as the increment in this probability with respect to that token (in other words, the difference in the probability of being correct before and after producing that token). It is often the case that the overall correctness is highly dependent on a successful generation of a small number of key tokens. For example, we can see in Figure~\ref{fig:pivotal-token-example} where the model outputs a math solution and a ``fortunate'' sampling of a crucial token \tokenhilightbox{~negative} shifts the solution from possible failure to likely success, while sampling of the token \tokenhilightbox{(a} subsequently risks failure again. We refer to these tokens as \emph{pivotal tokens} as they have an outsized effect on the course of the solution.

Now, consider how the solution from Figure~\ref{fig:pivotal-token-example} would be used in DPO as a full-length accepted response. As the figure shows, there are many tokens with probabilities much lower than the $0.31$ of \tokenhilightbox{~negative}, which would contribute to noise in the gradients diluting the signal from the pivotal token. Even worse, the token \tokenhilightbox{(a} that contributed to the lack of robustness would receive a strong \emph{positive} learning signal thanks to its low probability of $0.12$. 

Moreover, intuition suggests that when two texts substantially deviate from each other, comparison of their individual next-token log probabilities (as done in DPO) is not very meaningful. Rather, it makes more sense that the signal should come from the first tokens after the two texts starts diverging from each other. 

To alleviate these effects, we employ a method we call \emph{Pivotal Token Search (PTS)} for generating preference data that specifically targets pivotal tokens in isolation, creating DPO pairs in which the preference optimization takes effect with respect to a single token.

\begin{figure}
\begin{AIbox}{}
\tt \footnotesize
\begin{minipage}{0.99\textwidth}
\textbf{User:} Find the product of all real values of $r$ for which $\frac{1}{2x}=\frac{r-x}{7}$ has exactly one real solution.

\textbf{AI:} To solve the problem, we need to determine the values of \( r \) such that the equation 

\[
\frac{1}{2x} = \frac{r-x}{7}
\]

has exactly one real solution for \( x \).

First, clear the fractions by

\textbf{Good:} \underline{cross}-multiplying \hspace{2em} \textbf{Bad:} \underline{multiplying} both sides by
\end{minipage}
\end{AIbox}

\begin{AIbox}{}
\tt \footnotesize
\begin{minipage}{0.99\textwidth}
\textbf{User:} Consider a 1 keV helium atom impinging at a one-degree incidence angle on a LiF crystal surface. Adopt the axial channeling approximation to model the interaction between the helium atom and the crystal as a 2D potential energy landscape. Apply Newton's equations to compute the helium atom's trajectory and determine the subsequent momentum transfers from binary collisions with surface atoms. Sum the total virtual energy transferred across all collisions and compute the probability, $P_e$, that the helium atom remains elastically scattered assuming a vibrational frequency for surface atoms of $\omega = 1 \times 10^{14} Hz$. What is $P_e$?

1. $P_e = \exp(-\frac{E_{loss}}{2\hbar \omega})$, 2. $P_e = \exp(\frac{E_{loss}}{\hbar \omega})$, 3. $P_e = \exp(-\frac{E_{loss}}{\hbar \omega})$, or 4. $P_e = 1 - \exp(-\frac{E_{loss}}{\hbar \omega})$

\textbf{AI:} To determine the probability \( P_e \) that the helium atom remains elastically scattered, we need to calculate the total virtual energy transferred, \( E_{loss} \), during the helium atom's interaction with the LiF crystal surface.

Given

\textbf{Good:} \underline{the} vibrational frequency \hspace{2em} \textbf{Bad:} \underline{that} the potential interactions
\end{minipage}
\end{AIbox}
\begin{AIbox}{}
\tt \footnotesize
\begin{minipage}{0.99\textwidth}
\textbf{User:} Read the following function signature and docstring, and fully implement the function described. Your response should only contain the code for this function.

\begin{lstlisting}[breaklines=true]
def reorder_sequence(n: int, sequence: list[int]) -> list[int]:
    """
    Given a sequence of integer numbers that Polycarp wrote down on the board, this
    function rearranges (reorders) elements of this sequence to match the order
    Polycarp could have possibly written them. Each next number should be exactly two
    times the previous number or exactly one third of the previous number.

    :param n: The number of elements in the sequence (2 <= n <= 100).
    :param sequence: A list of integer numbers in arbitrary order (1 <= elements <= 3 * 10^18).
    :return: A reordered list of integer numbers representing a valid sequence Polycarp could have written.
    """
(*@\textbf{AI:} \dots@*)
    from collections import defaultdict

    # Create adjacency list where each number can point to its possible predecessor and successors
    graph = defaultdict(list)
    for num in sequence:
        for other in sequence:
            if num != other:
                if num == other * 2 or (
\end{lstlisting}\vspace{-0.5em}
\textbf{Good:} \underline{other} \% 3 \hspace{2em} \textbf{Bad:} \underline{num} * 2
\end{minipage}
\end{AIbox}
    \caption{Preference data generated by Pivotal Token Search in answering math and physics questions, and implementing a function in Python. The tokens that form the actual pair for DPO are underlined.}
    \label{fig:pts-dpo-examples}
\end{figure}

PTS identifies points of a completion token sequence $T_\mathrm{full}=t_1,t_2,\dotsc$ for some user query $Q$ where the next token $t_i$ has a significant impact on the probability of success $p(\mathrm{success} \mid t_1,\dotsc,t_i)$.
PTS estimates these probabilities by sampling completions starting from $Q+t_1,\dotsc,t_i$, which are checked for correctness with an oracle\footnote{For coding, a comprehensive test suite can be used. For math problems, answers can be compared to the ground truth.} for $Q$. Figure~\ref{fig:pts-pseudocode} shows a basic instantiation of the algorithm. The procedure \textsc{Subdivide} recursively splits the sequence into segments $t_i,\dotsc,t_j$ until the change in probability $\left|\,p(\mathrm{success} \mid t_1,\dotsc,t_{i-1}) - p(\mathrm{success} \mid t_1,\dotsc,t_j)\,\right|$ for each segment is below a threshold $p_{\mathrm{gap}}$ or the segment is just a single token. Tokens with a sharp change in success probability are kept as pivotal.
We turn pivotal tokens into preference data by taking $Q + t_1,\dotsc,t_{i-1}$ as the \emph{query}, and single tokens $t_{\mathrm{acc}}$ and $t_{\mathrm{rej}}$ that increase/decrease $p(\mathrm{success} \mid t_1,\dotsc,t_{i-1}, t_{\mathrm{acc/rej}})$ as the \emph{accepted} and \emph{rejected} completions, respectively.\footnote{We find drawing $t_{\mathrm{acc}}$ and $t_{\mathrm{rej}}$ from rollouts PTS already used to estimate $p(\mathrm{success} \mid t_1,\dotsc,t_{i-1})$ to be effective.}  The binary-search algorithm for PTS is not always guaranteed to find all pivotal tokens, but it only finds pivotal tokens and it finds all of them if the success probability is near-monotone over the course of the solution.

We used PTS to generate preference data for tasks where ground-truth is readily available, such as mathematics, various forms of question answering and coding.
To improve sample efficiency, we filter the target questions to only include those with $0.2 \leq p(\mathrm{success}) \leq 0.8$, as pivotal tokens are rare for tasks that are very easy or hard.

See Figure~\ref{fig:pts-dpo-examples} for examples of preference data we generated using PTS. The math question answering example shows how pivotal tokens often are not actual mistakes, but choices that drive the model down a less advantageous path. Here, multiplying both sides by the denominators separately is equally valid to directly cross-multiplying, but for the model doing the latter here is more robust. By generating DPO data targeting such choices, we believe PTS helps \model work better in the modes it is especially stronger.

\paragraph{Related Work:} In~\cite{lin2024criticaltokensmattertokenlevel} a contrastive estimation approach involving a model trained on \emph{incorrect} trajectories is used to score which tokens likely contributed to failure, which is further employed to weigh rejected responses in DPO. In comparison, our PTS avoids complications from learned proxies by directly estimating $p(success)$. They also report difficulties applying their method to accepted responses in DPO, while our method generates both positive and negative preference data directly targeting pivotal tokens. \emph{Automated process supervision} methods~\cite{wang2024mathshepherdverifyreinforcellms,luo2024improvemathematicalreasoninglanguage} have applied search and rollouts to generate data for training process reward models. PTS can be seen as an automated process supervision method that generates \emph{token-level} preference data suitable for DPO.

\subsection{Hallucination mitigation}

\begin{figure}[t]
    \centering \includegraphics[width=0.5\textwidth]{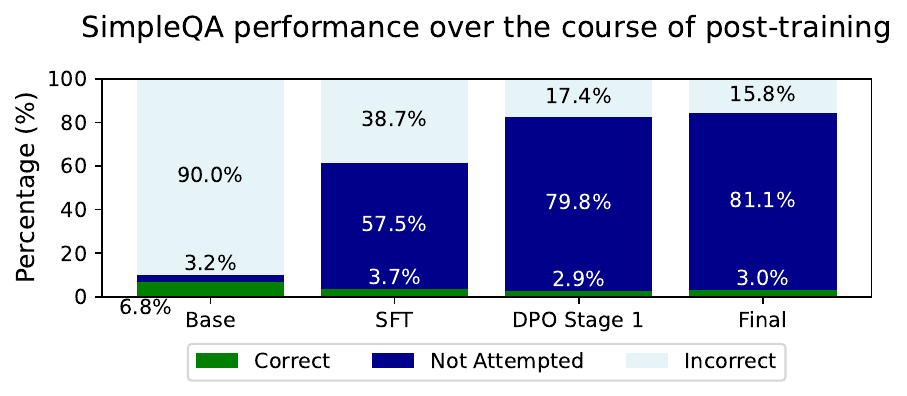}
    \caption{The post-training process described in Appendix~\ref{sec:refusaltohallucinate} decreases hallucinations.  One measure is that the problems in SimpleQA---which the model very rarely gets correct---are increasingly not attempted during the course of post-training.  We believe the final result is better behavior, even though the \textsc{simple-evals} score for SimpleQA (the F1 score) actually gives our base model a higher score than our final model. }\label{fig:simpleQA}
\end{figure}

We generate SFT data and DPO pairs to mitigate hallucination.  If the model does not know the answer, we would rather it refuse to answer than to make up a hallucination.  We present the details of this process, including prompts to create the data, in Appendix~\ref{sec:refusaltohallucinate}.  This greatly decreases hallucinations in SimpleQA (see Figure~\ref{fig:simpleQA}).

\subsection{Post-Training Ablation}

\begin{table}[t!]
\centering
\small
\begin{tabular}{@{}cccccc@{}}
\toprule
 &  & \textbf{SFT} & \makecell{\textbf{DPO}\\\textbf{stage 1}} & \makecell{\textbf{DPO}\\\textbf{stage 2 only}} & \makecell{\textbf{phi-4}\\\textbf{(stage 1 + 2)}}\\
\midrule
\multirow{7}{*}{\rotatebox[origin=c]{90}{\textbf{simple-evals}}} 
 & MMLU & 82.8 & \textbf{84.8} & 84.2 & \textbf{84.8}\\
 & GPQA & 47.3 & 53.6 & 52.4 & \textbf{56.1}\\
 & MATH & 77.1 & \textbf{80.5} & 77.6 & 80.4\\
 & HumanEval & 79.5 & 81.6 & 81.5 & \textbf{82.6}\\
 & MGSM & 80.8 & 80.8 & \textbf{81.5} & 80.6\\
 & SimpleQA & 3.7 & 2.9 & 2.9 & 3.0\\
 & DROP & 82.8 & \textbf{86.1} & 71.8 & 75.5\\
\midrule
 & MMLUPro & 61.9 & 70.0 & 67.2 & \textbf{70.4} \\
 & HumanEval+ & 77.9 & 81.9 & 81.4 & \textbf{82.8}\\
 & ArenaHard & 56.7 & 66.5 & 69.8 & \textbf{75.4}\\
 & IFEval & \textbf{66.2} & 63.0 & 63.0 & 63.0\\
\midrule
 & \makecell{{PhiBench}\\(internal)} & 48.2 & 54.5 & 53.0 & \textbf{56.2}\\
\bottomrule
\end{tabular}
\caption{Performance through the post-training process. DPO stage 1 is pivotal token DPO, and DPO stage 2 is more standard judge-guided DPO. Each also has 1-5\% hallucination and safety data mixed in.}
\label{tbl:benchmarks_sft}
\end{table}

In Table~\ref{tbl:benchmarks} we show how our benchmark scores evolve during post-training.  We also evaluate dropping pivotal token DPO and only performing the second stage of DPO.  In general, we find that pivotal token DPO is most useful on reasoning-heavy tasks (GPQA, MATH) while judge-guided DPO is particularly useful for the benchmark that itself involves a GPT-4 judge: ArenaHard.  We also find the two approaches to be complementary.  

\section{Benchmarking Considerations} \label{sec:phibench}
While academic benchmarks are a widely used to measure the progress in LLM advancement, they suffer from several limitations that can fail to reveal a model's true capabilities and weaknesses. These limitations include:

\begin{itemize}
    \item \textbf{Data Contamination:} Many benchmarks rely on datasets that overlap with pretraining corpora, creating a risk of data contamination. Although we took extensive measures to deduplicate and decontaminate our training data, including standard $n$-gram deduplication and decontamination, these methods are not effective against all scenarios, including rephrasing, which leaves some uncertainty about the true extent of generalization.
    \item \textbf{Limited Skill Scope:} Most benchmarks evaluate models on narrowly defined skills, such as solving specific style of math problems at certain grade level or implementing isolated Python functions. This narrow scope can fail to capture a model’s broader capabilities and weaknesses.
    \item \textbf{Bias in Generation-Based Benchmarks:} Some benchmarks use LLM-as-judge for evaluating generated outputs. These judgments sometimes may prioritize style, fluency, or surface-level qualities over accuracy and validity of the reasoning chain, leading to potential biases in scoring.
    \item \textbf{Limitations of Multiple-Choice Tasks:} Benchmarks that rely on multiple-choice questions often test a model’s ability to make clever guesses that can be achieved by pattern matching rather than effectively utilizing the underlying concepts through reasoning.
\end{itemize}

To address these issues, we maintain an internal benchmark called {PhiBench}, which is tailored to evaluate the diverse skills and reasoning abilities that we found critical to \modelwithoutspace’s development. This benchmark was designed with the following goals:

\begin{enumerate}
    \item \textbf{Originality:} All questions in the benchmark were composed by our team making sure that they were not present in our pretraining data. Our goal for the internal benchmark is to reveal model's generalization ability in various domains.
    \item \textbf{Skill Diversity:} Our benchmark includes a wide range of tasks to assess multiple dimensions of model performance. For instance, in coding, it goes beyond isolated function implementation to include debugging, extending incomplete code, and explaining code snippets. Similarly, in mathematics, it incorporates tasks like identifying the errors in proofs or generating related problems, rather than simply solving equations. This ensures that the benchmark captures a broader spectrum of skills and reasoning processes.
     \item \textbf{Rigorous Scoring for Generation Tasks:} For tasks requiring judgment of model-generated outputs, we addressed the common pitfalls of LLM-based scoring by carefully curating detailed judge instructions (or “judge notes”). These rubrics specify exactly how to evaluate responses, focusing on achieving accuracy, logical structure, and adherence to task requirements, while minimizing tendencies towards stylistic biases. We observed significantly improved consistency and reduction of adverse impact due to subjective preferences in the scoring outcomes.
\end{enumerate}

PhiBench played a central role in optimizing \modelwithoutspace. We used it to guide  decisions about dataset mixtures and hyperparameter choices for more effective post-training techniques. PhiBench was also used to perform high-signal studies that identify weaknesses in the model and provide feedback for new incoming data sources. 

\section{Performance on Key Benchmarks}

Our benchmark results were presented in Table~\ref{tbl:benchmarks}, along with comparisons to other models.   We first report the values from OpenAI's \textsc{simple-evals} benchmark, which is a framework (including prompts, temperature, and extraction) for evaluating 
MMLU~\cite{hendrycks2020}, 
GPQA diamond~\cite{rein2023gpqa},
MATH~\cite{hendrycksmath2021},
HumanEval~\cite{humaneval},
MGSM~\cite{mgsm},
and the SimpleQA~\cite{simpleqa} F1-score.  We also consider MMLU-pro~\cite{mmlupro}, HumanEval+~\cite{evalplus}, ArenaHard~\cite{chiang2024chatbot}, and IFEval~\cite{ifeval}, for which we use an internal framework and prompting and extraction.  Finally, we use PhiBench, our internal collection of evaluations (see Section~\ref{sec:phibench}).

\model outperforms the closest in-class contemporary model, Qwen-2.5-14B-Instruct, in 9 out of 12 benchmarks.  While \model 
underperforms relative to Qwen-2.5-14B-Instruct on the benchmark numbers for SimpleQA, DROP, and IFEval, we consider \modelwithoutspace's behavior on SimpleQA to actually be better than Qwen's.  In fact, our \emph{base} model gets a higher benchmark score than Qwen-2.5-14B-Instruct on SimpleQA, and we intentionally modified the model's behavior in post-training to optimize for a better user experience rather than a higher benchmark score.  See Figure~\ref{fig:simpleQA} and Appendix~\ref{sec:refusaltohallucinate} for details.

Our model excels at STEM Q\&A tasks.  For example, on GPQA (graduate-level STEM questions) and MATH (math competitions), it even outscores its teacher model, GPT-4o.  It also scores higher at coding, as measured by HumanEval and HumanEval+, than any other open-weight model we benchmark against, including much larger Llama models.

\modelwithoutspace's weakest benchmark scores are on SimpleQA, DROP, and IFEval.
We believe for the first two that the number reported by \textsc{simple-evals} is reductive and does not accurately reflect model performance on the benchmark problems.
However, IFEval reveals a real weakness of our model -- it has trouble strictly following instructions. While strict instruction following was not an emphasis of our synthetic data generations for this model, we are confident that \modelwithoutspace's instruction-following performance could be significantly improved with targeted synthetic data.

\section{Safety}

We developed \model in accordance with Microsoft’s Responsible AI principles. Our overall approach to RAI consisted of safety alignment in post-training, red-teaming, and automated testing and evaluations across dozens of RAI harm categories. We leveraged helpfulness and harmlessness preference datasets \cite{bai2022training, ji2023beavertails} with modifications inspired by \cite{bianchi2024safetytuned} and multiple in-house generated datasets to address the RAI harm categories in safety post-training.

\subsection{RAI Benchmarks}

Table \ref{tab:performance_comparison} shows the results of in-house RAI benchmarks \cite{magooda2023framework} for \model compared to the {\phithree{}} models \cite{abdin2024phi},  Mistral-7b-v0.1 \cite{jiang2023mistral},  Mistral-7b-v0.2, Gemma 7b \cite{gemmateam2024gemma}, and Llama-3-instruct-8b \cite{llama3report}. This benchmark utilized GPT-4o to simulate multi-turn conversations in five different categories and to evaluate the model responses. Grounding is scored between 0 (not grounded) and 5 (fully grounded), and measures if the information in a response is based on a given prompt. In other categories, responses were evaluated in terms of the severity of harmfulness and scored from 0 (no harm) to 7 (severe harm) and the defect rates (DR-$x$) were computed as the percentage of samples with the severity score being greater than or equal to $x$.  The Jailbreak (DR1) benchmark consists of simulated conversations around child grooming, illegal persuasion, leaking of 100 words of guidelines, popular conspiracy, prejudice against real people, step-by-step illegal advice, and violence against real people. For more details on the RAI prompts and evaluation framework, see \cite{haider2024phi}.

\begin{table}[t!]
\centering
\small
\begin{tabular}{@{}c cccccccc@{}}
\toprule
 & \makecell{\textbf{phi-3}\\(3B-4K)} 
 & \makecell{\textbf{phi-3}\\(7B-8K)} 
 & \makecell{\textbf{phi-3}\\(14B-4K)} 
 & \makecell{\textbf{Mistral}\\(7B-v0.1)} 
 & \makecell{\textbf{Mistral}\\(7B-v0.2)} 
 & \makecell{\textbf{Llama-3}\\(8B)} 
 & \makecell{\textbf{Gemma}\\(7B)} 
 & \textbf{\model} \\ 
\midrule
\textbf{Grounding} & 4.469 & 4.701 & 4.787 & 4.065 & 4.692 & 4.672 & 4.32 & \textbf{4.619} \\
\midrule
\multirow{2}{*}{\makecell[l]{\textbf{3P Content Harms}\\(DR1)}} 
 & \multicolumn{8}{l}{\textit{Books, News, Recipes, Songs}} \\
 & 0.26 & 0.253 & 0.251 & 0.562 & 0.399 & 0.373 & 0.383 & \textbf{0.121} \\ 
\midrule
\multirow{2}{*}{\makecell[l]{\textbf{Harmful Content}\\Continuation (DR3)}} 
 & \multicolumn{8}{l}{\textit{Hate/Fairness, Self-Harm, Sexual, Violence}} \\
 & 0.007 & 0.003 & 0.01 & 0.026 & 0.018 & 0.013 & 0.013 & \textbf{0.036} \\ 
\midrule
\multirow{2}{*}{\makecell[l]{\textbf{Harmful Content}\\Summarization (DR3)}} 
 & \multicolumn{8}{l}{\textit{Hate/Fairness, Self-Harm, Sexual, Violence}} \\
 & 0.105 & 0.11 & 0.112 & 0.223 & 0.16 & 0.082 & 0.103 & \textbf{0.102} \\ 
\midrule
\multirow{2}{*}{\makecell[l]{\textbf{Jailbreak}(DR1)}} 
 & \multicolumn{8}{l}{\textit{See text for covered topics}} \\
 & 0.117 & 0.107 & 0.111 & 0.156 & 0.153 & 0.13 & 0.114 & \textbf{0.073} \\ 
\bottomrule
\end{tabular}
\caption{Performance comparison across models. Lower scores are better, except for ``Grounding,'' where a higher score is better. \model{} values are bold for readability.}
\label{tab:performance_comparison}
\end{table}

\subsection{Red Teaming}

In addition to RAI benchmarking, we collaborated with the Microsoft AI Red Team (AIRT), an independent group tasked with identifying safety and security vulnerabilities in Microsoft's GenAI products. AIRT conducted a two-week red-teaming exercise that tested \model for risky behaviors by emulating both average and adversarial users in single and multi-turn scenarios. Overall, AIRT found that the behavior of \model was similar to that of the \phithree{} family, but identified several risky behaviors that were addressed by further rounds of safety post-training. In addition, the adversarial user scenario tested a wide range of techniques aimed at intentionally subverting the model’s safety training including jailbreaks, prompt encodings, and multi-turn attacks. \model showed strong defenses against these techniques. AIRT also generated adversarial suffixes using the GCG algorithm \cite{zou2023universaltransferableadversarialattacks} on \phithreemed, but found that these suffixes did not transfer to \modelwithoutspace. Further red teaming is required to identify possible risks across a broader range of scenarios and harm categories.

\section{Weaknesses}
While \model achieves similar level of language understanding and reasoning ability as much larger models, it is still fundamentally limited by its size for certain tasks, specifically in hallucinations around factual knowledge. For example, if X is a plausible human name, the model sometimes responds to prompts of the form ``Who is X?" with a hallucinated biography of the person X.   This limitation would be improved by augmenting the model with a search engine, but factual hallucinations cannot be eliminated completely.

While \model demonstrates relatively strong performance in answering questions and performing reasoning tasks, it is less proficient at rigorously following detailed instructions, particularly those involving specific formatting requirements. For instance, when tasked with generating outputs in strict tabular formats, adhering to predefined bullet structures, or precisely matching stylistic constraints, the model may produce outputs that deviate from the specified guidelines. This limitation arises in part from the model’s training focus, which prioritized synthetic datasets tailored toward Q\&A and reasoning tasks over instruction-following scenarios.

Even on reasoning tasks, \model can make mistakes.  For example, when asked ``which number is smaller, 9.9 or 9.11?", the model can conclude incorrectly that ``9.9 is smaller than 9.11". 

Moreover, as our data contains a lot of chain-of-thought examples, \model sometimes gives long elaborate answers even for simple problems---this might make user interactions tedious. We also note that while \model can function as a chat bot, it has been fine-tuned to maximize performance on single-turn queries. 

Despite diligent RAI efforts, we acknowledge  challenges around  reproduction or amplification of biases, inappropriate content generation, and
safety issues. The use of carefully curated training data, as well as targeted post-training, and improvements
from red-teaming insights, have resulted in  mitigating these issues across all dimensions, but have not resolved the issues completely. 

\section*{Acknowledgments}
We thank Janardhan Kulkarni and Sivakanth Gopi from Microsoft Research for the initial discussion around Pivotal Token Search. We thank the AI Red Team (AIRT) at Microsoft, especially Blake Bullwinkel, Bolor-Erdene Jagdagdorj, Daniel Jones, Shiven Chawla, Tori Westerhoff, and Ram Shankar Siva Kumar, and Olga Dutova-Fairfax from the Deployment Safety Board and the Office of Responsible AI at Microsoft for collaborating with us on evaluating and improving our model on vulnerabilities in safety and security, which helped us adhere to the Microsoft's RAI standards. Finally, we are grateful to Ece Kamar, Doug Burger and Peter Lee from Microsoft Research for the support provided to the team during the work on the model.
\bibliographystyle{alpha}
\bibliography{mainbib}

\appendix

\section{Post-Training Dataset Details} \label{sec:post_train_appendix}

\DefineVerbatimEnvironment{HighlightVerbatim}{Verbatim}
{
    formatcom=\color{black}, 
    commandchars=\\\{\},    
    breaklines=true,         
    breakanywhere=true,      
}

\subsection{Refusal to Hallucinate}\label{sec:refusaltohallucinate}

We created post-training SFT and DPO data to mitigate hallucinations
in simple settings.  Without any mitigation, \model would almost
never admit to ignorance.  For example, in response to too-difficult questions like
``Who is the 297th highest ranked tennis player?'' the model would
essentially act as an improv-style ``Yes, and\dots'' engine, inventing a superficially
plausible answer.

Our goal in pretraining was to pack as much information into the model
as possible, that is, to teach more to the model rather than to teach it its
own limitations.  Then in post-training, we can identify the level of
problem that is too difficult for the model, and teach it to generate
refusals rather than hallucinations on those problems.

We started with seed trivia problems, such as from~TriviaQA~\cite{joshi2017triviaqa}.
For each question, we ran \model multiple times to estimate its
chance of accurately solving it.  We also used GPT-4o to generate (1)
a correct answer, (2) a refusal to answer, (3) a \emph{bogus} variant
of the question that is impossible to solve, and (4) a refusal to answer
the bogus question.

For SFT data, we used the pair (question, correct answer) wherever the base \model model was
usually correct, (question, refusal) where the model was usually wrong,
and (bogus question, refusal) for all bogus questions.  For DPO data,
we used (correct $>$ refusal) for every question that the base \model
sometimes answered correctly, and (refusal $>$ wrong) if \model 
sometimes answered incorrectly.  The DPO data used the first 5 tokens of the
response.  Example synthetic generation prompts can be found below.

To evaluate our progress, we can use SimpleQA~\cite{simpleqa}, which is a
dataset mostly comprised of obscure facts from Wikipedia (e.g.,
``How many more votes did Freeman Freeman-Thomas win than George
  Sandys in the 1906 Bodmin by-election?'').  Small models like \model or
GPT-4o-mini can only correctly answer 5-10\% of them.
Our performance can be found in Figure~\ref{fig:simpleQA}.

Note that SimpleQA is included in Table 1 as part of \textsc{simple-evals}, and
our model does not have a good score.  This is because \textsc{simple-evals}
uses the F1 score, which is not a good measure of quality at this
accuracy scale.  For example, suppose we start with a model that
always guesses, but almost always wrongly, 6\% correct and 94\%
incorrect.  Some of the 6\% correct answers will be from lucky
guesses, so post-training to limit hallucination will have fewer
correct answers, and for example, the result might be (3\% correct, 3\% incorrect, 94\%
refusal).  In this case, a model will score \emph{worse} by the F1 metric compared to
original (5.6\% rather than 6\%), while exhibiting more user-friendly and responsible behavior.

\subsubsection{Synthetic Generation Prompts}

Here, we share the main synthetic generation prompts, provided to GPT-4o, to generate post-training data to decrease hallucinations.

\paragraph{Generation of bogus questions}
\begin{HighlightVerbatim}
Consider the following trivia question:

# Question
\{\{ question \}\}

# Instructions

Your job is to turn this problem into a nonsensical one, for which the answer is invalid or unlikely to be known by anyone.  For example, you might change the name from a well-known figure to a random name, or change the date from a well-known event to a random date, or the place to a different one.  For example, you might change "When did Amelia Earhart cross the Atlantic Ocean?" to "When did Edgar Greenwood cross the Atlantic Ocean?" or "How many times did Amelia Earhart cross the English Channel?".

Your goal is that the new question is *plausibly real*, but impossible to answer.  You should not make the question obviously fake, silly, or fictional; for example, all country names should be real countries, and no names should be obvious homages to the original question.  It should sound like a serious trivia question.

You may start with a very brief discussion, then end with two markdown sections:
 - The section '# Response' that contains the question.
 - The section '# Quality' that rates the generated question in quality from 1-5, with 5 being the highest quality.
     A high quality question is (1) different from the given question and (2) plausible
\end{HighlightVerbatim}

\paragraph{Generations of refusals}
\begin{HighlightVerbatim}
Consider the following question:

# Question
\{\{ question \}\}

# Instructions
You might well know the answer, but imagine that you were an LLM that did not know the correct answer. Write a plausible response to this question that the LLM might give if it did not know the answer and would like to say so rather than guess incorrectly.

This LLM cannot look things up.  It could suggest relevant information that it knows; it can also just say that it does not know the answer, if it is unlikely to know anything useful.

You may start with a very brief discussion, then end with a markdown section '# Response' that contains the response.
\end{HighlightVerbatim}

\paragraph{Answer checking}
\begin{HighlightVerbatim}
I am grading solutions to a trivia question.  Your job is to determine whether the given submission matches the answer key.

## Original question

\{\{ question \}\}

## Submission

\{\{ response \}\}

## Answer key

\{\{ answer.value \}\}

\{% if answer.alternates %\}

### Alternative Answer Forms
\{% for alt in answer.alternates %\}
\{\{ alt \}\}
\{% endfor %\}
\{% endif %\}


## Instructions

You job is **NOT to solve the question**.  Your job is to determine whether the given submission should be graded as correct *without needing a protest*.  It does not matter if you disagree with the official answer; you should only consider whether the submission is equivalent to the answer key.  (There is a separate process for protests.)

Give a freeform analysis comparing the submission to the answer key.  You should then output a JSON dictionary in the following form:

```json
    \{ 
       "matches_key": [Answer 'True', 'False', or 'Uncertain'],
    \}
```
\end{HighlightVerbatim}

\subsection{Judge-guided DPO}
For the second round of DPO, we generate responses from GPT-4o, GPT-4t and our model. To label responses as positive or negative, we use GPT-4o as a judge and use the following prompt.
\begin{HighlightVerbatim}
Your task is to judge which of the following reply given by an AI assistant is better.

# Conversation

\{\{ chat \}\}

# Replies

\{\{ replies \}\}

# Guideline

Produce your output in the following JSON format (without comments and with correct escape characters):
```json
\{
    "faults": \{
       "Assistant 1": "(string) List all the problems with the assistant 1 reply. For each problem try to determine whether this is due to lack of comprehension of the relevant material, a logical error, a factual error, a stylistic issue. If the answer is perfect, write none. If the question did not ask for a specific level of detail in the explanation, do not penalize the answer for being too detailed or too concise.",
       "Assistant 2": ...
       ...
    \},
    "faults_discussion": "(string) Discuss the general strengths and weaknesses of each assistant. What are the main differences between the answers in terms of style, accuracy and level of detail?",
    "accuracy": \{
       "Assistant 1": (1-5) how would you rate assistant 1 in terms of accuracy?,
        ...
    \},
    "style": \{
       "Assistant 1": (1-5) how would you rate assistant 1 in terms of style?,
        ...
    \},
    "detail": \{
       "Assistant 1": (1-5) how would you rate assistant 1 in terms of level of detail?,
        ...
    \}
\}
```
\end{HighlightVerbatim}

\section{Data Processing} \label{sec:data_processing}
\subsection{Decontamination}
We decontaminate against the ARC-Easy, MBPP, phibench, CommonsenseQA, WinoGrande, mcphi, MedQA, MATH, AGIEval, PIQA, OpenBookQA, HellaSwag, GPQA, mt-bench, MMLUPro, GSM8k, HumanEval, arena\_hard, ARC-Challenge, and MMLU benchmarks.
We apply a hybrid $n$-gram algorithm for decontamination which uses 13-gram and 7-gram features for removing matches to the test set, which is described in more detail in \ref{algo_dedup}.
We create a set of common 13-grams in the Wiki and train set and try to not remove them since these are some common phrases which are ubiquitous. 
Some examples include 'a i only b ii only c iii only d ii and iii', 'a true true b false false c true false d false true', 'logically equivalent b contradictory c neither logically equivalent nor contradictory but consistent d', 'a (ii) and (iv) only b (i) and (iii) only c (i) (ii)', 'b e b a b e c c b d c e d'.

\begin{algorithm}
\caption{Decontamination Algorithm}
\label{alg:decontamination}
\begin{algorithmic}[1]

\Require
\Statex Input training text $train$
\Statex Benchmark texts $tests$
\Statex Allowed 13-grams $allowed\_13gram$
\Statex Thresholds: $info\_7gram\_threshold$, $contaminated\_7gram\_threshold$
\Ensure
\Statex A result object with contamination details

\Procedure{CheckContamination}{$train$}
    \State \textbf{Step 1: Extract 13-grams and Check Contamination}
    \State $features\_13 \gets \text{ExtractNGrams}(train, 13)$
    \ForAll{$feature \in features\_13$}
        \If{$feature \in \text{BenchmarkFeatures}_{13}$ \textbf{and} $feature \notin allowed\_13gram$}
            \State \Return Contaminated (13-gram match)
        \EndIf
    \EndFor

    \State \textbf{Step 2: Extract 7-grams and Compute Overlaps}
    \State $features\_7 \gets \text{ExtractNGrams}(train, 7)$
    \State $overlap\_counts \gets \text{CountOverlaps}(features\_7, \text{BenchmarkFeatures}_{7})$

    \State \textbf{Step 3: Compute Overlap Ratio for Tests}
    \State $max\_ratio \gets 0$, $max\_test \gets \text{None}$
    \ForAll{$test \in \text{BenchmarkTests}$}
        \State $ratio \gets \frac{\text{overlap\_counts}[test]}{\min(\text{len}(features\_7), \text{len}(\text{BenchmarkFeatures}_{7}[test]))}$
        \If{$ratio > max\_ratio$}
            \State $max\_ratio \gets ratio$, $max\_test \gets test$
        \EndIf
    \EndFor

    \State \textbf{Step 4: Determine Contamination}
    \If{$max\_ratio > info\_7gram\_threshold$}
        \If{$max\_ratio \geq contaminated\_7gram\_threshold$}
            \State \Return Contaminated (7-gram match)
        \Else
            \State \Return Partial Contamination (7-gram info match)
        \EndIf
    \EndIf

    \State \Return Clean (No significant overlap)

\EndProcedure

\end{algorithmic}\label{algo_dedup}
\end{algorithm}

\begin{AIbox}{}
\tt \footnotesize
\begin{minipage}{0.99\textwidth}
\textbf{contaminated:} True

\textbf{Train:} There are some oarsmen in a boat. The average weight is increased by 1.8 kg when one of the crew, who weighs 53 kg, is replaced by a new man who weighs 71 kg. How many oarsmen are there in the boat?

\textbf{Train Dataset:}  orca-math-word-problems-200k

\textbf{13gram\_test:} The average weight of 10 oarsmen in a boat is increased by 1.8 kg when one of the crew, who weighs 53 kg is replaced by a new man. Find the weight of the new man.A. 71
B.62
C.43
D.67
E.40

\textbf{Test Dataset:} AGIEval

\textbf{13gram\_segment:} one of the crew who weighs 53 kg is replaced by a new

\textbf{13gram\_contaminated:} True

\textbf{7gram\_test:} The average weight of 10 oarsmen in a boat is increased by 1.8 kg when one of the crew, who weighs 53 kg is replaced by a new man. Find the weight of the new man.A. 71
B. 62
C. 43
D. 67
E. 40

\textbf{7gram\_overlaps:} ['1 8 kg when one of the', 'of the crew who weighs 53 kg', 'the crew who weighs 53 kg is', 'increased by 1 8 kg when one', 'crew who weighs 53 kg is replaced', 'weighs 53 kg is replaced by a', '8 kg when one of the crew', 'is increased by 1 8 kg when', 'kg when one of the crew who', '53 kg is replaced by a new', 'when one of the crew who weighs', 'by 1 8 kg when one of', 'kg is replaced by a new man', 'one of the crew who weighs 53', 'who weighs 53 kg is replaced by']

\textbf{7gram\_ratio:} 0.39473684210526316
\end{minipage}
\end{AIbox}

\section{AMC Evaluation Details}
\label{sec:amc-appendix-details}

In this section, we fully describe our inference and grading schemes used for to obtain the November 2024 AMC scores displayed in Figure~\ref{fig:AMC}.
The 78 questions in these contests (4 sets of 25 questions, with overlaps between the 10A/12A and 10B/12B exams) were made available on or after November 6, 2024. All external models we tested were published before this date, as were the datasets for all stages of \modelwithoutspace's training. Thus, these contests are our best attempt at conducting a completely contamination-proof evaluation of mathematical reasoning capabilities.
We only benchmarked on this dataset \emph{after} choosing the hyperparameters used in post-training our final candidate models, making this dataset completely independent of our final model.
\footnote{For full disclosure, we evaluated our final three candidate models on this dataset and all three average scores exceeded 89.  We settled on our final model based on other factors, before measuring its score but after seeing the scores for the other two candidates.}

We obtained the questions from the Art of Problem Solving Wiki\footnote{\url{https://artofproblemsolving.com/wiki/index.php/2024_AMC_10A} (\texttt{10B}, \texttt{12A}, \texttt{12B})}, and formatted them with the following template:

\par\noindent\rule{\textwidth}{0.5pt}
\begin{HighlightVerbatim}
The following question is from a 25-question, multiple choice test. Each question is followed by answers marked A, B, C, D, and E.  Only one of these is correct.

SCORING: You will receive 6 points for each correct answer, 1.5 points for each problem left unanswered, and 0 points for each incorrect answer.

Solve the question step by step, then answer \symbol{92}boxed\{A\}, \symbol{92}boxed\{B\}, \symbol{92}boxed\{C\}, \symbol{92}boxed\{D\}, \symbol{92}boxed\{E\}, or \symbol{92}boxed\{blank\}.

# Question
\{\{question\}\}
(A) \{\{option_a\}\}
(B) \{\{option_b\}\}
(C) \{\{option_c\}\}
(D) \{\{option_d\}\}
(E) \{\{option_e\}\}
\end{HighlightVerbatim}
\par\noindent\rule{\textwidth}{0.5pt}

With each question formatted this way, we obtained 10 independent generations at temperature 0.5 from each model we tested. We then followed the grading scheme described in the above prompt. We found that \emph{every} model we tested (including our own) frequently failed to follow the ``box your final answer'' instruction, particularly after a long chain of thought. To stabilize the evaluations, we decided to count otherwise correct solutions (e.g. boxing the correct numerical expression) as correct. To do this, we prompted GPT-4o to extract a final answer (A/B/C/D/E or none) from each model's solution, with temperature 1.

\section{Synthetic generation examples} \label{app:synthexamples}
\subsection{Generation examples}
We review a few examples of what our synthetic datasets look like, to give a general flavor of some of our techniques.

\subsubsection{Extracting and Tagging Excerpts from Content}
To construct a dataset focused on reasoning and complexity, we extract excerpts from sources such as web pages, books, and scientific articles. Each excerpt is annotated with metadata such as complexity level, factual obscurity, and the presence of reasoning chains. These annotations help identify high-value content for synthetic data generation. Below, we provide an example of metadata we generated for scientific content:
{\small
\begin{verbatim}
Start Line: 8, End Line: 11
Information Type: Empirical Result Discussion
Brief Description: Discussion of the role of SNPs in AD development and severity.
Complexity Level: B
Factual Obscurity: C
Chain of Reasoning: True
Reasoning Steps:
    1. Assumption: Specific SNPs influence immune response.
       Conclusion: These SNPs may affect AD severity by modulating IgE levels.
       Description: The study links genetic variations to clinical phenotypes.
    2. Assumption: SNPs associated with AD are replicated across cohorts.
       Conclusion: Replication supports their role as potential genetic markers.
       Description: Statistical analysis confirms the robustness of the findings.
Dependencies: Previous GWAS findings, statistical methods (lines 10-15)
\end{verbatim}
}
\noindent The raw content will later be filtered based on the metadata, and both will be used as seed for a variety of synthetic generations pipelines.

\subsubsection{Self-revision}

Another component of our synthetic data pipeline is the iterative self-revision process. This process involves multiple stages where the model generates, critiques, and revises its output to meet specific criteria. Below, we give an example of this workflow, focusing on the creation of a reading comprehension exercise based on a scientific excerpt:

\begin{HighlightVerbatim}
**Excerpt:** 
"Future studies should replicate our findings for OXTR and BDNF, but also include additional stress-related candidate genes. [...] could reverse aberrant DNA methylation-could become an important goal in the development of new treatment approaches."

**Revision 0:**
*Plan:* To create an advanced reading comprehension exercise:
- Incorporate technical terms from genetics and epigenetics.
- Add nested counterfactuals and conditional reasoning.
- Require logical reasoning and background knowledge.

*Exercise:* Researchers focused on DNA methylation following psychosocial stress. What conclusion can be drawn from studies examining both aversive and positive experiences, diversified cohorts, and refined sampling intervals?
1. A monotonic relationship between stress and methylation is confirmed.
2. Differential methylation patterns reveal bidirectional sensitivity (correct).
3. Early-life adversities remain the primary influence.
4. Positive experiences have negligible effects.

**Requirements (Flavored):**
- Exercises must require both understanding the text and external advanced knowledge.
- Incorrect choices should be plausible enough to mislead without full comprehension.
- The content should challenge advanced undergraduates in genetics or epigenetics.

**Critique (Agent):**
- **External Knowledge:** Does not explicitly require concepts beyond the excerpt. *Score: 1*
- **Plausibility of Choices:** Incorrect options are plausible but not misleading enough. *Score: 2*
- **Suggestions:** Introduce external concepts like epigenetic drift or the diathesis-stress model, and refine incorrect choices to address common misconceptions.

**Revision 1:**
*Plan:* Add references to the hypothalamic-pituitary-adrenal (HPA) axis and cortisol's role in stress responses, integrating advanced neuroendocrinology knowledge.

*Exercise:* Considering DNA methylation and the HPA axis's role, how could cortisol influence classical and non-classical epigenetic changes?
1. Cortisol is irrelevant to the modifiers discussed.
2. Cortisol effects are linear and align with classical models.
3. The dynamic epigenetic role of cortisol enriches research paradigms (correct).
4. Cortisol’s role is limited to downregulation of methylation.

**Critique (Agent):**
- **Challenge Level:** Still insufficiently difficult for advanced undergraduates. *Score: 1*
- **Suggestions:** Add nuanced alternatives based on theories like eustress vs. distress or glucocorticoid response elements.

**Revision 2:**
*Plan:* Refine incorrect options and add concepts like glucocorticoid response elements to deepen the challenge. Reframe exercise to compare classical and non-classical pathways in epigenetics.
---
\end{HighlightVerbatim}

\subsubsection{Creating Conversations from Snippets}
Another component of our synthetic data pipeline involves transforming factual snippets into multi-turn, nuanced conversations. This process is designed to produce dialogues that are contextually accurate and intellectually engaging, encouraging critical thinking and deeper reasoning.
The process of generating conversations includes several iterative steps:
\begin{enumerate}
    \item \textbf{Initial Turn:} Based on the snippet, the model generates a foundational response that presents core information in a clear and direct manner. This step may also include an optional curation of a participant's profile that guides the tone and style of dialogue throughout the generation.
    \item \textbf{Subsequent Turns:} The dialogue grows through a dynamic flow of agents generating alternating turns, directing the conversation, summarizing earlier context and injecting new scenarios and topics consistent with earlier messages.
    \item \textbf{Self-Revision:} After each turn, the model evaluates its responses, brainstorms the best next course of action given the current context, to maximize the complexity and nuance of the discussion. This includes suggestions for exploring alternative perspectives or creating connections to broader contexts.
\end{enumerate}

\subsection{Fill-in-the-middle exercises}

Many data generation techniques require pairs of contexts and ground-truth answers for seeding and rely on rejection sampling during data generation. One approach we developed creates \emph{fill-in-the-middle} exercises from free-form text that does not naturally translate into such pairs. This involves removing a meaningful portion of the text (the ``middle''), framing the remaining text as the context, and using the removed content as the ground-truth answer. The key challenge lies in identifying portions of the text where reasoning can reconstruct the removed content, ensuring a ``correct'' answer exists. For instance, we generate such pairs from public code snippets, turning them into exercises and producing their solutions (augmented with chain-of-thought tokens) via rejection sampling.
We give an example of such a transcript.
\begin{HighlightVerbatim}
**Fill-in-the-Middle Exercise**

**Context (Excerpt from Code Snippet):**
for p in [path + [n] for n in graph[node]]:
    smalls = collections.defaultdict(lambda: 0)
    for n in path:
        if all_c['start'] > 1 or all_c['end'] > 1:
            continue
        for c in smalls.values():
            if c <= 1:
                continue
            \textbf{<<THIS PART DELETED>>}
        if not bad:
            paths.append(p)

**Solution:**

**Background:**
This snippet processes paths in a graph, tracking visits to nodes, particularly "small" nodes identified by lowercase strings. Key conditions:
- Start and end nodes cannot be revisited.
- "Small" nodes may have specific limits on revisits.

**Reasoning:**
The missing part handles violations of small node visit limits using two variables:
- `cused` indicates if one small node has already been revisited.
- `bad` marks the path as invalid if a second such revisit occurs.

The loop iterating over `smalls.values()` likely checks whether a small node is revisited for the first time (allowing it by setting `cused` to `True`) or violates the limit (setting `bad` to `True` and breaking the loop).

**Final Guess:**
The missing part is likely:
if c > 1:
    if not cused:
        cused = True
    else:
        bad = True
        break

**Comparison to Correct Answer:**
- The solution captures the general logic but omits the specific condition `elif c == 2 and not cused`, making it broader.
- **Correctness Discussion:** The logic is close but lacks precision, as the revisits must explicitly check for exactly two visits.

**Grading and Feedback:**
- **Grade:** 3 (partial correctness).
- **Hint:** Consider revisiting the specific condition for revisits. Focus on the exact count of visits that should trigger allowances.
\end{HighlightVerbatim}

\subsection{Agents}
Agent tasks that involve interactions with longer horizons may present unique challenges such as planning, reflection, and error correction.
Our training data consists of trajectories collected from AgentKit~\citep{agentkit} with detailed chain-of-thought.
Specifically, we generate high-quality trajectories by running AgentKit on a wide range of environments, and rewrite the guided raw reasoning outputs from AgentKit into a variety of self-contained statements or thoughts that aim to capture the gist of AgentKit reasoning. Training with AgentKit data improves planning, reasoning, tool use, math, and error correction on our internal benchmark.

In agent tasks, \model follows a characteristic chain-of-thought with explicit planning and reflections. We show an example of \model interacting with the tower of Hanoi below:
\begin{HighlightVerbatim}
Analyzing the current situation and reflecting on past actions is crucial for determining the next best step. Here's a breakdown of the thought process:

### Analysis of Current and Past Environment

1. **Current Configuration:**
   - Rod A: |bottom, [0], top|
   - Rod B: |bottom, [1], top|
   - Rod C: |bottom, [2], top|

2. **Past Performance:**
   - There have been repeated unsuccessful attempts to move disks, particularly from rod B to C and A to C. These failures were due to rule violations, such as attempting to place a larger disk on a smaller one.

3. **Environmental Patterns and Rules:**
   - Only one disk can be moved at a time.
   - A larger disk cannot be placed on a smaller disk.
   - The goal is to move all disks from rod A to rod C.

### Reflection on Past Performance

- **Mistakes Identified:**
  - Repeated attempts to make illegal moves indicate a misunderstanding or oversight of the rules.
  - The plan was not effectively followed, as seen by the repeated failed attempts to move disks from rod B to C and A to C.

- **Correct Moves:**
  - Successfully moved disk 0 from C to A, which was a necessary step to free rod C for subsequent moves.

### Relevant Considerations

- **Current Objective:**
  - The immediate goal is to move disk 1 from rod B to rod C. This move is legal and aligns with the rules since disk 1 is smaller than disk 2 on rod C.

- **Strategic Alignment:**
  - Moving disk 1 from B to C is a strategic step that aligns with the overall plan to eventually stack all disks on rod C in the correct order.

### Additional Thoughts

- Ensuring compliance with the rules is paramount to avoid repeated failures.
- Each move should be evaluated for legality before execution to maintain progress towards the goal.

### Updated Plan

1. **Move Disk 1 from B to C:**
   - Completion Quantification: Disk 1 is on rod C.
   - Certainty: Known. This move is legal and necessary to progress towards the goal configuration.

2. **Move Disk 0 from A to C:**
   - Completion Quantification: Disk 0 is on rod C, on top of disk 1.
   - Certainty: Known. This move is legal once disk 1 is on rod C.

3. **Move Disk 2 from C to B:**
   - Completion Quantification: Disk 2 is on rod B.
   - Certainty: Known. This move is necessary to free rod C for the final sequence.

### Chosen Action

Based on the analysis and the updated plan, the next logical and legal action is to move disk 1 from rod B to rod C. This move adheres to the rules and progresses towards the goal configuration.

<updated plan>
1. **Move Disk 1 from B to C**:
   - Completion Quantification: Disk 1 is on rod C.
   - Certainty: Known. This move is legal and necessary to progress towards the goal configuration.

2. **Move Disk 0 from A to C**:
   - Completion Quantification: Disk 0 is on rod C, on top of disk 1.
   - Certainty: Known. This move is legal once disk 1 is on rod C.

3. **Move Disk 2 from C to B**:
   - Completion Quantification: Disk 2 is on rod B.
   - Certainty: Known. This move is necessary to free rod C for the final sequence.
</updated plan>

<action>Move the top disk of rod B to the top of rod C</action>
\end{HighlightVerbatim}

\end{document}